%% file: main.tex
\documentclass{article}

\PassOptionsToPackage{numbers, compress}{natbib}
 \usepackage[preprint]{neurips_2026}

\input{preamble}
\usepackage[utf8]{inputenc} % allow utf-8 input
\usepackage[T1]{fontenc}    % use 8-bit T1 fonts
\usepackage{hyperref}       % hyperlinks
\usepackage{url}            % simple URL typesetting
\usepackage{booktabs}       % professional-quality tables
\usepackage{amsfonts}       % blackboard math symbols
\usepackage{nicefrac}       % compact symbols for 1/2, etc.
\usepackage{microtype}      % microtypography
\usepackage{xcolor}         % colors
\usepackage{array}
\usepackage{multirow}
\usepackage{graphicx}
\usepackage{placeins}
\usepackage{subcaption}
\usepackage[dvipsnames]{xcolor}

\title{Anatomy-Preserving Unpaired Medical Image Translation via Shared Latent Anchoring}

\author{%
  \textbf{Zhilin Zou$^{1}$ \quad
  Hieu Le$^{2}$ \quad
  Saumya Gupta$^{1}$} \\
  \textbf{Jingyi Xu$^{1}$ \quad
  Prateek Prasanna$^{1}$ \quad
  Chao Chen$^{1}$} \\
  $^{1}$Stony Brook University, NY, USA \\
  $^{2}$University of North Carolina at Charlotte, NC, USA \\
  \texttt{zhizou@cs.stonybrook.edu} \quad
  \texttt{chao.chen.1@stonybrook.edu}
}
\begin{document}
\newcommand{\parag}[1]{\vspace{-4mm}\paragraph{#1}}
\maketitle

\input{sections/abstract}
\input{sections/introduction}
\input{sections/related}

\input{sections/method}

\input{sections/experiment}
\input{sections/conclusion}
\clearpage
\FloatBarrier

\section*{Acknowledgments}
This research was partially supported by grants NSF CCF-2144901, NIH R01NS143143 and R01CA297843.

{
    \small
    \bibliographystyle{IEEEtran}
    \bibliography{reference}
}

%%%%%%%%%%%%%%%%%%%%%%%%%%%%%%%%%%%%%%%%%%%%%%%%%%%%%%%%%%%%
\clearpage
\appendix
\input{sections/appendix}

% \section{Technical appendices and supplementary material}
% Technical appendices with additional results, figures, graphs, and proofs may be submitted with the paper submission before the full submission deadline (see above). You can upload a ZIP file for videos or code, but do not upload a separate PDF file for the appendix. There is no page limit for the technical appendices. 

% Note: Think of the appendix as ``optional reading'' for reviewers. The paper must be able to stand alone without the appendix; for example, adding critical experiments that support the main claims to an appendix is inappropriate. 

%%%%%%%%%%%%%%%%%%%%%%%%%%%%%%%%%%%%%%%%%%%%%%%%%%%%%%%%%%%%
% \clearpage
% \newpage
% \input{checklist.tex}

\end{document}

%% file: preamble.tex
\usepackage{colortbl} % for cell colors
\usepackage{multirow} % for merging rows
\usepackage{xcolor}   % for color definitions
\usepackage{array}    % for table alignment
\usepackage{subcaption}
\usepackage{graphicx}
\usepackage{booktabs}
\usepackage{algorithm}
\usepackage{amssymb}
\usepackage{wrapfig}
\usepackage{algpseudocode}
\usepackage{fancyhdr}
\usepackage{mathtools}
\usepackage{bm}

\newcommand{\myparagraph}[1]{\smallskip\noindent\textbf{#1}}

\newcommand{\best}[1]{\textbf{#1}}
\usepackage{amsmath}

%% file: sections/abstract.tex
\begin{abstract}
    Unpaired medical image-to-image translation requires synthesizing target-modality images from source-modality images without aligned supervision. Existing methods learn a direct source-to-target mapping that entangles structure preservation and appearance generation within a single model, often distorting clinically meaningful anatomy in the translated output. We argue that these two objectives should be decoupled entirely: anatomical structure should be captured independently of generation, and the generator should operate only on a frozen structural representation. To this end, we propose a two-stage framework for unpaired medical image translation. In the first stage, we learn a shared structure space by training a structure encoder-decoder on source images, target images, and structural masks jointly, using mask reconstruction to anchor the latent space to modality-agnostic anatomy. For source images without mask annotations, a target-domain memory bank provides soft structural supervision via prototype retrieval. In the second stage, a conditional flow model learns to render target-domain appearance under fixed structural conditions. This ensures the generator inherits target appearance statistics and can focus entirely on structural conditioning rather than learning appearance from scratch. We validate our framework on OCT-to-OCTA synthesis and brain MR-to-CT translation. Our method outperforms all unpaired baselines on both tasks and, despite requiring no paired supervision, also surpasses paired baselines, improving PSNR by up to 11.2\% and SSIM by up to 26.4\% on OCT-to-OCTA, and achieving consistent gains across whole-image, soft-tissue, and bone regions on MR-to-CT. The project repository is available on
\href{https://github.com/RNZZL/Anatomy-Preserving-Unpaired-Medical-Image-Translation-via-Shared-Latent-Anchoring}{GitHub}.
    % In the second stage, the structure encoder-decoder is frozen, and a conditional flow model learns to render target-domain appearance under the fixed structural conditions. This initialization ensures the generator inherits target appearance statistics and can focus entirely on structural conditioning rather than learning appearance from scratch.
\end{abstract}

%% file: sections/introduction.tex
\section{Introduction}

% 1st paragraph: What is the problem // Why is the problem important?
%    don't discuss related work too much here
% talk about the problem: multi-modality importance in diagnostics. difficulty is in taking repeated scans, patient discomfort etc

% 2nd paragraph: Why is the problem difficult?
%    related work (learning-based) // Give the intuition of your idea, get the reader hooked

% 3rd paragraph: What is your key idea? Broad!! Disentanglement part. You can point to the teaser figure.

% 4th paragraph: Give the important technical details.
%    I often have to move aspects mentioned in the Method section here

% 5th paragraph: How do you evaluate your method? What are the most impressive results?
%    you can already explain here why you did not evaluate on some dataset the reviewer could think of.

% 6th paragraph: A summary of your contribution (possibly as a bullet list).
%    you can also add a paragraph after the 5th paragraph to explain what your method is *not* if you think the reviewer can see relations with unrelated previous work

% 7th paragraph: Maybe a short overview of the paper. 

% Converting medical images from one modality to another while preserving clinically meaningful structure is a central task in medical image-to-image translation. 
Multi-modal medical imaging is important in clinical diagnosis because different imaging modalities provide complementary anatomical, functional, and physiological information for disease assessment~\cite{syndiff}. However, acquiring multi-modal images remains challenging due to factors such as cost, acquisition time, scanner availability, patient motion, and protocol inconsistency~\cite{syndiff}. For example, optical coherence tomography (OCT) captures high-resolution retinal structure but not blood flow, whereas OCT angiography (OCTA) provides vascular information but requires additional hardware and software support~\cite{chen2025mutri}, which limits its availability. Such trade-offs between modality utility and accessibility are common across clinical imaging settings. Medical image-to-image translation, therefore, offers a practical solution by synthesizing missing target modalities from available source modalities while preserving source anatomical structures. A key challenge, however, is that most existing translation methods require \textit{paired} source-target data during training~\cite{isola2017image,zhou2023denoising,zhang2025structure}, where images from different modalities are spatially aligned and correspond to the same subject, anatomy, and acquisition setting. 
Such paired multi-modal data are difficult to obtain in practice, especially when modalities differ in acquisition protocols, availability, scan timing, field of view, or patient compliance. 
This motivates \textit{unpaired} translation, where source- and target-domain images are available as separate collections without one-to-one correspondence.
% A key challenge of the translation task, however, is that most existing methods require \textit{paired} data during training~\cite{isola2017image, zhou2023denoising, zhang2025structure}, where images from different modalities are spatially aligned and correspond to the same subject, anatomy, and acquisition setting. Such paired multi-modal data is often difficult to obtain in practice, especially when modalities differ in acquisition protocols, availability, or patient compliance. This motivates \textit{unpaired} translation, where images from multiple modalities are available but not aligned across them. 

% Further, the training data for this problem is often \textit{unpaired} since modalities may differ in acquisition protocols, availability, scan timing, field of view, and patient compliance. As a result, source-domain images and target-domain images are often available only as separate collections, without one-to-one correspondence.

%%%%%%%%%%%%%Figure 1%%%%%%%%%%%%%%%%
\input{figures/teaser}
%%%%%%%%%%%%%%%%%%%%%%%%%%%%%%%%%%%

% \hl{Medical image translation is not simply an appearance-transfer problem. Although modalities differ in contrast, texture, intensity distribution, and imaging response, they are often constrained by the same patient-specific anatomy. This shared anatomy is the bridge across domains: OCT and OCTA are tied by retinal geometry, vessel organization, and the foveal avascular zone (FAZ), while MR and CT share brain tissue, ventricles, and skull-related structures. A translated image is therefore clinically meaningful only if it changes the modality-specific appearance while preserving the underlying anatomy. Existing unpaired translation methods, however, usually learn a direct source-to-target mapping through adversarial matching, cycle consistency, contrastive patch learning, or diffusion-based conditional generation~\cite{syndiff,zhu2017unpaired,park2020contrastive}. These objectives encourage target-domain realism, but they do not explicitly specify which anatomical information must remain invariant. This entangles structure preservation with appearance synthesis, and without paired supervision, the model can produce target-like images that still contain missing vessels, distorted FAZ boundaries, shifted ventricles, or hallucinated tissue patterns. As illustrated in Fig.~\ref{fig:teaser}, direct translation can produce target-like appearance while still distorting clinically meaningful structures. This motivates an explicit formulation that defines the cross-domain structure to be preserved before target-domain appearance is generated.}

Medical image translation is not simply an appearance-transfer problem. Although modalities differ in contrast, texture, intensity distribution, and imaging response, the translated image should preserve the anatomy of the input source image. These modality-shared anatomical structures provide a natural basis for cross-domain translation: OCT and OCTA are tied by vessel organization and the foveal avascular zone (FAZ), while magnetic resonance imaging (MR) and computed tomography (CT) share brain tissue, ventricles, and skull-related structures. A translated image is therefore clinically meaningful only if it changes modality-specific appearance while preserving the underlying anatomy. Existing unpaired translation methods usually learn a direct source-to-target mapping through adversarial matching, cycle consistency, contrastive patch learning, or diffusion-based conditional generation~\cite{syndiff,zhu2017unpaired,park2020contrastive}. 
These objectives encourage target-domain realism, but they do not explicitly specify which anatomical information should remain invariant. 
As a result, structure preservation and appearance synthesis are entangled, and without source-target correspondence, the model may produce target-like images with missing vessels, distorted FAZ boundaries, or hallucinated tissue patterns, as illustrated in Fig.~\ref{fig:teaser}. 
This motivates an explicit formulation that defines the cross-domain structure to be preserved before target-domain appearance is generated.

To this end, we ground the translation process in the underlying anatomical structure. 
Instead of learning a direct source-to-target mapping, we formulate unpaired medical image translation as structure-conditioned target-domain generation. We first construct a cross-domain shared structure latent space that specifies the anatomy to be preserved across modalities. A structure encoder-decoder maps source images, target images, and available structural masks into this shared space and reconstructs structural maps, encouraging the latent representation to capture modality-agnostic anatomy rather than modality-specific appearance. We then decouple structure extraction from target-domain appearance synthesis by using the learned structure space as fixed anatomical guidance for generation.

Learning such a shared structure space under unpaired supervision is challenging. On the one hand, accurately registered cross-domain image pairs are difficult to acquire. On the other hand, obtaining reliable structural annotations for both modalities is also unrealistic in many medical settings, because different modalities reveal different structures with different reliability. For example, bone-related structures are easier to define in CT than in MR, while vascular structures are more directly visible in OCTA than in OCT. Therefore, we avoid assuming accurate source-domain structural annotations. 
Instead, we propose a Structural Memory Bank (SMB), which stores target-domain structural masks and their latent representations as prototypes. 
For each source image, SMB retrieves structurally similar target-domain prototypes and aggregates them into a soft structural target, enabling supervision of source-derived structure latents using only target-domain structural annotations. We then freeze the learned structure encoder-decoder and use its structural representations to condition a target-modality flow model~\cite{lipman2022flow}. Initialized from an unconditional target-domain prior, the flow model focuses on rendering target appearance under fixed anatomical guidance, thereby separating structure preservation from appearance generation.

% In Stage II, the frozen structure encoder-decoder serves as a structural prior to condition a target-domain flow model. Target images supply realistic appearance supervision, while source images provide structural conditions for cross-domain synthesis. Because the flow model is initialized from an unconditional target-domain prior, it begins with the target appearance distribution and learns only to follow the structural guidance, allowing clean decoupling of the two objectives.

% To achieve this, we propose a two-stage framework for unpaired medical image translation. We disentangle structure-preservation from appearance-generation and do each stage-wise. In the first stage, we learn a shared structure space with a structure encoder-decoder. The encoder maps source images, target images, and available structural masks into spatial and global structural representations, while the decoder reconstructs structure maps from these representations. This design encourages the latent space to capture modality-agnostic anatomical layouts rather than modality-specific appearance. In the second stage, the learned structure encoder-decoder is frozen as a structure prior and used to condition a target-domain flow model. Target-domain images are used to learn realistic target appearance, while source-domain images provide structural conditions for cross-domain synthesis. As a result, the generator is guided by an explicit structural representation learned before image synthesis, allowing it to focus on target-modality appearance generation.

We evaluate the proposed framework on two representative medical image translation tasks: OCT to OCTA and Brain MR to CT translations. These tasks involve different anatomies, imaging mechanisms, and appearance transformations while sharing the same goal: preserving the source-domain structure while generating target-domain appearance. Experiments show that explicitly modeling the shared structure to be preserved improves both image fidelity and structural consistency compared with competing baselines, supporting the effectiveness of structure-conditioned generation for unpaired medical image translation. 

In summary, our contributions are as follows:
\begin{enumerate}
\item \textbf{Structure-conditioned formulation.} We formulate unpaired medical image translation as structure-conditioned target-domain generation. This explicitly defines modality-shared anatomical structure as the information to be preserved and target-domain appearance as the information to be generated.

% \cc{This is technically not true. If it is only an disentanglement problem, you are extracting the structure and appearance components, which together will reconstruct the original image.}

\item \textbf{Shared structure latent modeling with SMB.} We construct a cross-domain shared structure latent space using a structure encoder-decoder, and propose a Structural Memory Bank (SMB). SMB stores target-domain structural prototypes and provides soft supervision for source-derived structure latents, supporting shared structure learning through target-domain structural supervision.
% \pp{why are you mentioning partial annotations here?}.

\item \textbf{Comprehensive experimental validation.} We evaluate our framework on OCT-to-OCTA and brain MR-to-CT translation,
demonstrating improved image fidelity and structural consistency over
competing paired and unpaired baselines. Held-out-center testing,
independent vessel evaluation, downstream classification, and a blinded
reader study further assess generalization and preservation of
clinically relevant information.

\end{enumerate}

%% file: figures/teaser.tex
% \begin{wrapfigure}{r}{0.48\textwidth}
%     \vspace{-10pt}
%     \centering
%     \includegraphics[width=0.46\textwidth]{figures/teaser.png}

%     \caption{
%     Conceptual comparison between conventional unpaired translation and our shared structure-conditioned generation. }
%     \label{fig:teaser}
%     \vspace{-10pt}
% \end{wrapfigure}

\begin{wrapfigure}[19]{r}{0.45\textwidth}
    \vspace{-17pt}
    \centering
    \includegraphics[width=0.40\textwidth]{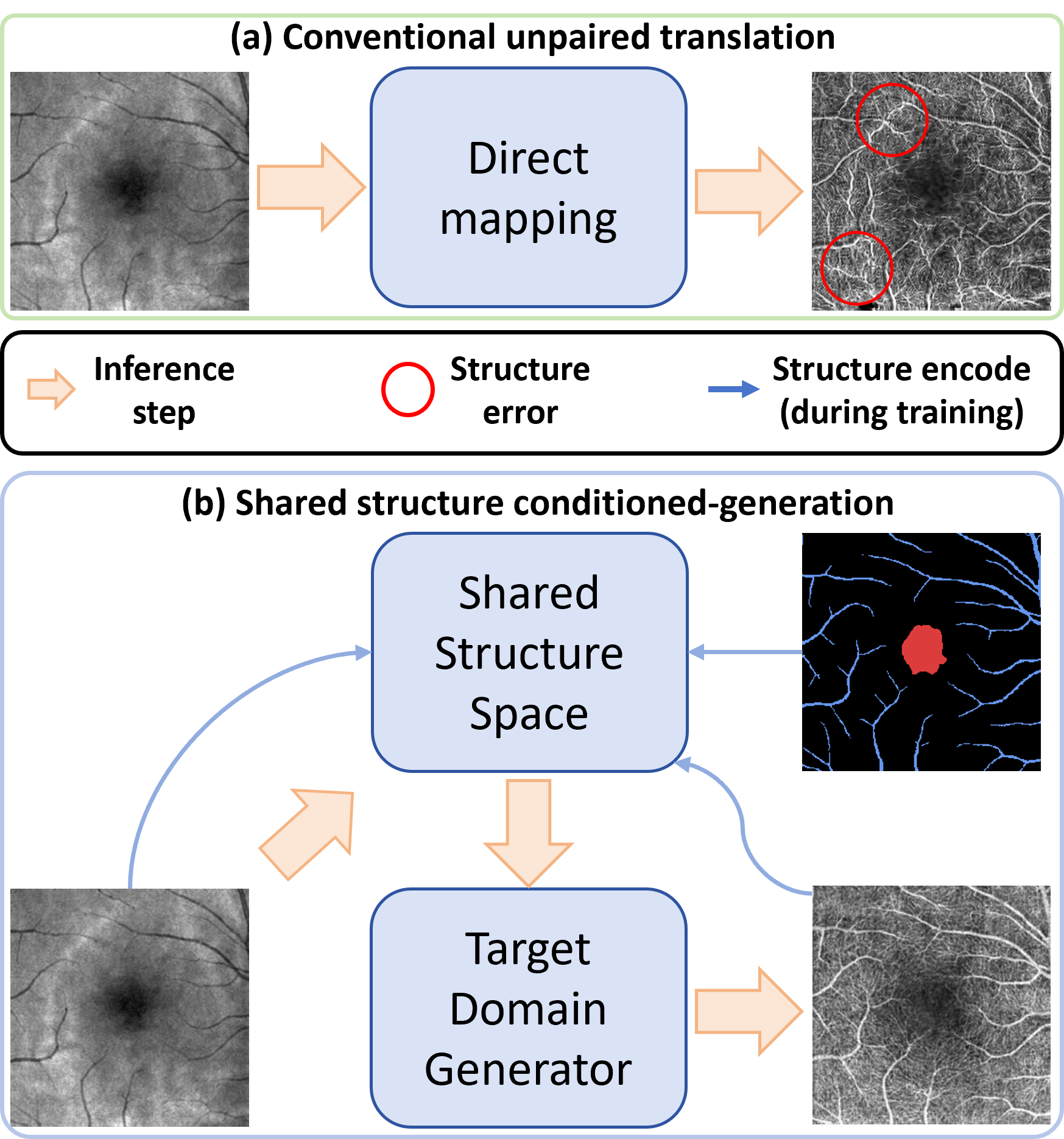}
    \vspace{-6pt}
    \caption{
    Conventional unpaired translation vs. our shared structure-conditioned generation.}
    \label{fig:teaser}
    \vspace{-16pt}
\end{wrapfigure}

%% file: sections/related.tex
\section{Related work}
\parag{Image-to-image translation.}
Image-to-image (I2I) translation has been widely studied for cross-modality medical image synthesis, aiming to infer a target modality from an acquired source modality. 
Supervised methods require paired data, such as Pix2pix~\cite{isola2017image}, diffusion-bridge models~\cite{zhou2023denoising}, MuTri for paired 3D OCT-to-OCTA synthesis~\cite{chen2025mutri}, and MISA-LDM for missing-modality MRI synthesis~\cite{zhang2025structure}. 
Despite their effectiveness, these methods depend on accurately registered multimodal data, which are difficult to collect at scale. Unpaired I2I translation reduces this requirement by learning cross-domain mappings from unaligned image collections~\cite{zhu2017unpaired,park2020contrastive,huang2018multimodal,fu2019geometry,benaim2017one,jung2022exploring,wang2021instance,zheng2021spatially}. 
Representative methods include CycleGAN~\cite{zhu2017unpaired}, which uses cycle consistency, CUT~\cite{park2020contrastive}, which adopts patch-wise contrastive learning, and MUNIT~\cite{huang2018multimodal}, which disentangles content and style. 
Recent approaches further explore probabilistic formulations, such as UNSB~\cite{kim2024unpaired} with Neural Schr\"odinger Bridges, and SynDiff~\cite{syndiff} with cycle-consistent adversarial diffusion for medical image translation. 
Although these methods reduce the need for paired data, they still largely learn translation as a direct domain mapping, where structure preservation and appearance generation are entangled. 
In the related setting of unpaired image restoration,
DDSB~\cite{yi2026degradation} incorporates degradation-aware optimal
transport and dynamic transport consistency into a Schr\"odinger bridge.
Our method instead uses a frozen shared structure representation to decouple anatomical structure modeling from target-domain appearance rendering.
% In contrast, our method uses a frozen shared structure representation to decouple anatomical structure modeling from target-domain appearance rendering.

\parag{Conditional generative models for medical image synthesis.}
We focus on conditional generative models for medical image synthesis. 
Beyond unconditional diffusion models such as DDPMs~\cite{ho2020denoising} and transport-based models such as flow matching~\cite{lipman2022flow}, recent methods improve controllability using anatomical, pathological, or physical conditions. 
LeFusion~\cite{zhang2024lefusion} and spine MRI inpainting~\cite{hansen2024inpainting} use lesion or pathology masks for structure-aware synthesis and editing, while adaptive diffusion priors~\cite{ahmed2025pinn}, Dose-Aware Diffusion~\cite{xie2024dose}, and Poisson-flow generative models~\cite{hein2024ppfm} incorporate acquisition or physics constraints for MRI, PET, and CT reconstruction. 
Topology-aware methods further introduce persistent-homology or multi-scale topological constraints for synthetic masks and cellular layouts~\cite{gupta2024topodiffusionnet,xu2025topocellgen}. 
More recently, SynFlow~\cite{asadi2026fms} uses mask-conditioned flow matching to generate topology-preserving crack and vessel images from binary masks. 
Overall, these studies demonstrate the value of ROI/mask conditioning, physics guidance, topology-aware constraints, and flow-based generation, but generally require image-condition correspondence during training, which can be difficult to obtain in medical settings.

%% file: sections/method.tex
\section{Method}
\label{sec:method}

We consider unpaired medical image translation from a source domain $\mathcal{X}_A$ to a target domain $\mathcal{X}_B$, where no image-level correspondence is available across domains. As both domains depict the same anatomical organ or region, they contain comparable anatomical structures, although their appearance (such as intensity and noise)  can differ substantially due to modality- or protocol-specific characteristics. We assume access to target-domain image--mask pairs $(x_B, m_B)$, where $m_B$ denotes the structural mask of image $x_B$. In the source domain, only images $x_A \sim \mathcal{X}_A$ are required while coarse source masks $m_A$ (obtained using threshold-based methods) may be used when available. Given a source image $x_A$, the goal is to generate the corresponding target-domain image $\hat{x}_B$ that preserves the anatomical structure of $x_A$ while matching the appearance distribution of $\mathcal{X}_B$.

% Directly learning an unpaired mapping from $\mathcal{X}_A$ to $\mathcal{X}_B$ can entangle structure preservation with target-appearance generation, which may lead to anatomical distortion or unrealistic synthesis. 
% We formulate unpaired translation as shared structure latent-conditioned target-domain generation. 
We formulate unpaired translation as latent structure-conditioned target-domain generation, and we tackle each of them in separate stages. We first briefly review the flow matching~\cite{lipman2022flow} formulation used for target-domain generation. In Stage I, described in Sec.~\ref{sec:stageI}, we learn a shared structure encoder-decoder ($E_{\omega}, D_{\phi}$). The encoder maps source images, target images, target-domain masks, and optional source-domain masks into a common structure space $\mathcal{Z}_{\mathrm{str}}$, while the decoder reconstructs structural maps, providing an explicit structure prior shared across domains. In Stage II, described in Sec.~\ref{sec:stageII}, we freeze $E_{\omega}$ and $D_{\phi}$ and train a conditional flow model to generate a target-domain image $x_B$ from the learned structural conditions $s_u\in \mathcal{Z}_{\mathrm{str}}$. Sec.~\ref{sec:traininfer} summarizes the inference procedure. We provide an overview in  Fig.~\ref{fig:pipeline}.

% With this initialization, the generator inherits target-domain appearance statistics, allowing conditional training to use the extracted anatomical structure to guide target-domain image generation. 

 % \saum{i think we also need to describe the data setting correctly. For target, we have both image and mask, while for source, we only have image and optional mask right?}

% The rest of this section is organized as follows. We briefly discuss flow models in Sec.~\ref{sec:flow}, followed by details of Stage I in Sec.~\ref{sec:stageI} and Stage II in Sec.~\ref{sec:stageII}. Finally, we tie our framework together to provide the training procedure and inference steps in Sec.~\ref{sec:traininfer}.

%%%%%%%%%%%%%%%%%%%%%%%%%%%%%%%%%%%%%
\input{figures/pipeline}

%%%%%%%%%%%%%%%%%%%%%%%%%%%%%%%%%%%%%

% \subsection{Background: Flow Matching}
% \label{sec:flow}
% Flow matching~\cite{lipman2022flow} learns a time-dependent velocity field that transports samples from a Gaussian distribution to the desired data distribution. In this work, we use pixel-space flow matching. Given a target-domain image $x\sim \mathcal{X}$, a Gaussian noise sample $\epsilon \sim \mathcal{N}(0, I)$, and time step $t \in [0,1]$, the linear interpolation path is defined as: $x_t = (1 - t)\epsilon + t x$.
% Along this path, if we differentiate with respect to $t$, we obtain the target velocity: $v_t = x - \epsilon$. 
% An unconditional flow model $v_{\theta}(x_t, t)$ is trained to predict transportation velocity by minimizing:
% \begin{equation}
% \mathcal{L}_{\mathrm{flow}}
% =
% \mathbb{E}_{\epsilon,t}
% \left[
% \left\| v_{\theta}(x_t,t) - v_t \right\|_2^2
% \right].
% \end{equation}
% During inference, generation starts from Gaussian noise and proceeds by solving the ordinary differential equation (ODE):
% \begin{equation}
% \frac{d x_t}{d t} = v_{\theta}(x_t,t),
% \end{equation}
% from $t = 0$ to $t = 1$, producing a sample from the learned target-domain distribution. For conditional generation, the velocity field is augmented with a condition. 
% Given a training image $x$ and its condition $s$, the conditional flow model is trained by:
% \begin{equation}
% \mathcal{L}_{\mathrm{condition}}(x,s)
% =
% \mathbb{E}_{\epsilon,t}
% \left[
% \left\|
% v_{\theta}(x_t,t \mid s)
% -
% v_t
% \right\|_2^2
% \right].
% \end{equation}

\parag{Background: Flow Matching.}
\label{sec:flow}

Flow matching~\cite{lipman2022flow} learns a velocity field that maps Gaussian noise to the data distribution. For $x\sim\mathcal{X}$, $\epsilon\sim\mathcal{N}(0,I)$, and $t\in[0,1]$, we define $x_t=(1-t)\epsilon+tx$, giving target velocity $v_t=x-\epsilon$. The unconditional and conditional objectives are
\begin{equation}
\label{eq:condition}
\mathcal{L}_{\mathrm{flow}}
=
\mathbb{E}_{\epsilon,t}
\left[
\left\|
v_{\theta}(x_t,t)-v_t
\right\|_2^2
\right],
\quad
\mathcal{L}_{\mathrm{cond}}
=
\mathbb{E}_{\epsilon,t}
\left[
\left\|
v_{\theta}(x_t,t\mid s)-v_t
\right\|_2^2
\right].
\end{equation}
At inference, samples are generated by solving $\frac{dx_t}{dt}=v_{\theta}(x_t,t)$ from $t=0$ to $t=1$.

\subsection{Stage I: Shared Structure Space Learning}
\label{sec:stageI}
Stage I learns an encoder-decoder to construct a modality-agnostic latent structure space. Given an input $u$, the structure encoder $E_{\omega}$ produces a latent spatial feature $s_u = E_{\omega}(u)$ 
where $s_u \in \mathcal{Z}_{\mathrm{str}}$, and $u$ can be either source image $x_A$, target image $x_B$, or target mask $m_B$. The structure decoder $D_{\phi}$ reconstructs a multi-class structural map, $\hat{m}_u = D_{\phi}(s_u)$. Since images and masks differ in their low-level form, we use input-specific adapters before the shared encoder layers. These adapters absorb low-level input differences, while the shared layers enforce a common latent structure representation. 
Stage I uses different supervision depending on the input type. Target mask inputs $m_B$ anchor the latent space through reconstruction, target images $x_B$ are supervised by their structural masks $m_B$, and source images $x_A$ are guided by memory-bank-retrieved target-domain prototypes when reliable source-domain masks are unavailable. When a coarse source mask $m_A$ is available, it provides additional weak decoding supervision. These complementary objectives encourage all inputs to be encoded into a shared structure space $\mathcal{Z}_{\mathrm{str}}$, as detailed below.

\textbf{Mask decoding}. To anchor $\mathcal{Z}_{\mathrm{str}}$ to anatomical structure, we supervise the decoder output $\hat{m}$ with an available structural mask $m$, using a mask decoding loss:
\begin{equation}
\mathcal{L}_{\mathrm{dec}}(\hat{m},m)
=
\mathrm{CE}(\hat{m}, m)
+
\frac{1}{C-1}
\sum_{c=1}^{C-1}
\left(
1 - \mathrm{Dice}(\hat{m}^{c}, m^{c})
\right),
\end{equation}
where $\mathrm{CE}(\cdot,\cdot)$ denotes the cross-entropy loss, $\mathrm{Dice}(\cdot,\cdot)$ denotes the class-wise Dice coefficient, $C$ is the number of structure classes, and the background class is excluded from the Dice term. 
This loss is applied to different inputs depending on the available supervision: $m_B$ defines the mask-based structure space by self-reconstruction, $x_B$ is supervised by its corresponding annotation $m_B$, and $x_A$ is supervised by the coarse mask $m_A$ only when it is available.

\textbf{Image-mask latent alignment}. To encourage the target-domain image and corresponding mask to share a canonical structure space, we align their latent features via a latent consistency loss:
\begin{equation}
\mathcal{L}_{\mathrm{align}}
=
\left\|
{s}_B - {s}_m
\right\|_1,
\quad
s_B = E_{\omega}(x_B), \quad s_m = E_{\omega}(m_B),
\end{equation}
where ${s}_B$ and ${s}_m$ denote the spatial features of the image-induced and mask-induced latent representations, respectively. 

\textbf{Soft prototype structure latent alignment}. It is essential for $(E_\omega, D_\phi)$ to encode structural information from source images $x_A$, as this is the only input available during inference time. However, source-domain images often lack reliable mask annotations. To provide structural supervision in this setting, we propose a Structural Memory Bank (SMB), 
% \saum{i don't like this name at all. It is super confusing and very long to read. Also, `Retrieve' automatically implies retrieval based on some similarity measure. Can we just name it as `Structural Memory Bank (SMB)'?}
which stores target masks, target images, and their corresponding image latent representations as a prototype set $\mathcal{P}$. For each source image, SMB retrieves structurally similar target prototypes and aggregates them into a soft structural latent $\bar{s}$, providing pseudo structural supervision without requiring source-domain masks. 
% The memory bank stores target-domain masks and their corresponding latent representations as a prototype set $\mathcal{P}$. For each source image, MRMB retrieves structurally similar target prototypes and aggregates their latent representations into a soft prototype structural latent $\bar{s}$, which serves as pseudo structural supervision for the source input.
% to learn a soft prototype structural latent $\bar{s}$. The prototype set $\mathcal{P}$ stores target masks and their latent representations: 
\begin{equation} 
\mathcal{P} = \{(x_B^j,m_B^j, s_B^j)\}_{j=1}^{N}, \quad s_B^j = E_{\omega}(x_B^j), 
\end{equation} 
where $N$ is the number of stored target domain samples. Given a source image $x_A$, we predict its structure map $\hat{m}_A = D_{\phi}(E_{\omega}(x_A))$, then compare $\hat{m}_A$ with target masks $\{m_B^j\}_{j=1}^{N}$ in the bank and retrieve the top-$K$ most similar prototypes (similarity measured via Dice). For each retrieved prototype $\{s_B^j\}_{j=1}^{K}$, we compute a similarity-based weight $\alpha_j$, and the retrieved prototypes are then aggregated into the soft prototype structural latent $\bar{s}$:
\begin{equation}
\bar{s}
=
\sum_{j =1}^K
\alpha_j s_B^j,
\quad
\alpha_j
=
\frac{
\exp\left(\mathrm{Dice}(\hat{m}_A,m_B^j)^{\gamma}/\tau\right)
}{
\sum_{\ell \in K}
\exp\left(\mathrm{Dice}(\hat{m}_A,m_B^{\ell})^{\gamma}/\tau\right)
},  
\end{equation}
where $\gamma$ controls the sharpness of similarity scores, and $\tau$ is a temperature parameter. 
The source-derived latent is then aligned with this soft latent using an $L_1$ loss:
\begin{equation}
\mathcal{L}_{\mathrm{bank}}
=
\left\|
E_{\omega}(x_A) - \bar{s}
\right\|_1.
\end{equation}
This bank-based supervision encourages source structures to lie close to the target structural distribution without requiring paired image supervision.

\textbf{Cycle consistency}. To further suppress modality-specific appearance in the structure latent space $\mathcal{Z}_{\mathrm{str}}$, we introduce an asymmetric latent-cycle consistency constraint. Specifically, we decode the input-derived latent into a structural mask and then re-encode this mask back into the latent space. The original input latent is encouraged to match the re-encoded structural latent.
\begin{equation}
\mathcal{L}_{\mathrm{cycle}}
=
\left\|
s_u - \operatorname{sg}(\tilde{s}_u)
\right\|_1,
\quad
s_u = E_{\omega}(u),
\quad
\tilde{s}_u = E_{\omega}(D_{\phi}(s_u)).
\end{equation}
where $\operatorname{sg}(\cdot)$ denotes the stop-gradient operation, and $u \in \{x_A, x_B\}$. Therefore, the cycle loss only updates the first encoded latent $s_u$, while the re-encoded latent $\tilde{s}_u$ is treated as fixed. This asymmetric image-to-mask-to-latent cycle encourages $s_u$ to retain information that can be consistently expressed through a structural map, while suppressing modality-specific appearance cues.

% We further apply two auxiliary regularization items: view augmentation to improve robustness of the structure representation, and mask-space adversarial regularization to encourage predicted structures to match the target mask distribution.

% The main Stage I objective is therefore written as:
% \begin{equation}
% \mathcal{L}_{\mathrm{stageI}}
% =
% \lambda_{\mathrm{dec}}\mathcal{L}_{\mathrm{dec}}
% +
% \lambda_{\mathrm{align}}\mathcal{L}_{\mathrm{align}}
% +
% \lambda_{\mathrm{bank}}\mathcal{L}_{\mathrm{bank}}
% +
% \lambda_{\mathrm{cycle}}\mathcal{L}_{\mathrm{cycle}}.
% \end{equation}
% \saum{need to mention Ldec twice, one for m and one for xB}
% The reconstruction loss defines the mask-based structure space, the alignment loss pulls annotated image representations into this space, the bank loss aligns source-derived structures with the target structural distribution, and the cycle loss suppresses modality-specific appearance information by enforcing image-to-structure-to-latent consistency. 
% After Stage I, the structure encoder $E_{\omega}$ and decoder $D_{\phi}$ are frozen and used to provide latent structural conditions for the target-domain flow generator.

\textbf{Overall Objective in Stage I}. To summarize, the Stage I objective depends on the type of input. For a target structural mask $m_B$, we use only the reconstruction loss:
\begin{equation}
\mathcal{L}_{m_B}
=
\lambda_{\mathrm{dec}}
\mathcal{L}_{\mathrm{dec}}
\big(D_{\phi}(E_{\omega}(m_B)), m_B\big).
\end{equation}
For a target image $x_B$ with its structural annotation $m_B$, we use reconstruction, image-mask latent alignment, and latent cycle consistency:
\begin{equation}
\mathcal{L}_{x_B}
=
\lambda_{\mathrm{dec}}
\mathcal{L}_{\mathrm{dec}}
\big(D_{\phi}(E_{\omega}(x_B)), m_B\big)
+
\lambda_{\mathrm{align}}
\mathcal{L}_{\mathrm{align}}(x_B, m_B)
+
\lambda_{\mathrm{cycle}}
\mathcal{L}_{\mathrm{cycle}}(x_B).
\end{equation}
For a source image $x_A$, accurate annotations are unavailable. We therefore use the memory-bank loss to pull its structure latent toward the target structural distribution, together with latent cycle consistency. 
If an optional coarse source mask $m_A$ is available, we also include a decode loss:
\begin{equation}
\mathcal{L}_{x_A}
=
\lambda_{\mathrm{dec}}^{A}
\mathcal{L}_{\mathrm{dec}}
\big(D_{\phi}(E_{\omega}(x_A)), m_A\big)
+
\lambda_{\mathrm{bank}}
\mathcal{L}_{\mathrm{bank}}(x_A)
+
\lambda_{\mathrm{cycle}}
\mathcal{L}_{\mathrm{cycle}}(x_A),
\end{equation}
where $\lambda_{\mathrm{dec}}^{A}=\lambda_{\mathrm{dec}}$ when the optional coarse source mask $m_A$ is available, and $\lambda_{\mathrm{dec}}^{A}=0$ otherwise. The overall Stage-I objective is optimized by applying the corresponding loss to each type of input:
\begin{equation}
\mathcal{L}_{\mathrm{stageI}}
=
\mathcal{L}_{m_B}
+
\mathcal{L}_{x_B}
+
\mathcal{L}_{x_A}.
\end{equation}
The reconstruction losses anchor the structure space using available masks, the alignment loss pulls annotated target image representations into this space, the bank loss aligns source-derived structures with the target structural distribution, and the cycle loss suppresses modality-specific appearance information through image-to-mask-to-latent consistency. 

\subsection{Stage II: Structure-Conditioned Target Generation}
\label{sec:stageII}
In Stage II, we freeze the encoder-decoder $(E_{\omega}, D_{\phi})$ learned in Stage I, and train a target-domain flow generator conditioned on structural representations from $\mathcal{Z}_{\mathrm{str}}$. This decouples the two objectives: Stage I extracts structure, while Stage II learns to generate target-domain appearance. 
Given a latent structural condition $s\in\mathcal{Z}_{\mathrm{str}}$, the conditional velocity field is defined as $v_{\theta}(x_t,t\mid s)$, where $s$ provides structural guidance. 
The generator is optimized to render target-domain appearance while respecting the structural condition from the frozen encoder. As in Stage I, Stage II's training objective varies with the input type; we describe these as self-branch and cross-branch below. 

\myparagraph{Target-domain self-branch.} This branch learns to render target appearance under reliable structural supervision. Given a target image $x_B$ and its mask $m_B$, we obtain the flow model condition as $s_B = E_{\omega}(m_B)$. The conditional flow model is then trained to reconstruct $x_B$ under this condition using the conditional flow matching objective (from Eq.~\ref{eq:condition}):
%Given a target image $x_B$ and its mask $m_B$, we obtain the target structural condition by encoding the mask with the frozen structure encoder $s_B = E_{\omega}(m_B)$. The conditional flow model is then trained to generate $x_B$ under this condition using the conditional flow matching objective:
\begin{equation}
\mathcal{L}_{\mathrm{self}}
=
\mathcal{L}_{\mathrm{cond}}(x_B,s_B).
\end{equation}
This branch provides in-domain supervision for the generator, teaching it to render target-domain appearance when the corresponding structure is provided as condition.

\myparagraph{Source-to-target cross-branch.} This branch teaches the flow model to render target appearance when the condition is derived from the source image. Given a source image $x_A$, we first predict its structural map using the frozen encoder-decoder as $\hat{m}_A = D_{\phi}(E_{\omega}(x_A))$. To make the condition closer to the mask-defined structure space, we do not directly use the image-derived latent representation $E_{\omega}(x_A)$ 
% \saum{technically $\hat{m}_A$ is not latent rep ; maybe you should talk about it as $s_1 = E_{\omega}(x_A)$ and then $s_2 = E_{\omega}(D_{\phi}(E_{\omega}(x_A)))$. Also, you claim $s_1$ has appearance-leakage, but did you do experiment to confirm that $s_2$ performs better than $s_1$?} 
to condition the flow model. Instead, the predicted structural map is \textit{re-encoded} into the shared structure space $s_A = E_{\omega}(\hat{m}_A)$. This image-to-mask-to-condition pathway ensures that the generator is conditioned mainly on source anatomy rather than source-domain appearance. Since no corresponding target image is available for $x_A$, we reuse the target-domain structure bank introduced in Stage I to construct a structure-matched soft target image. Based on the mask similarity between $\hat{m}_A$ and target masks in the bank, we retrieve the top-$K$ target images and use the similarity weights $\alpha_j$ to construct a structure-matched soft target image. 
The cross-domain training objective is then defined as
\begin{equation}
\tilde{x}_B
=
\sum_{j \in K}
\alpha_j x_B^j,
\quad
\mathcal{L}_{\mathrm{cross}}
=
\mathcal{L}_{\mathrm{cond}}(\tilde{x}_B, s_A).
\end{equation}
The soft target image $\tilde{x}_B$ is not treated as ground-truth supervision for $x_A$. Instead, it provides a target-domain appearance reference with a similar structure, enabling the generator to learn target-style rendering for source-derived structure conditions. This facilitates unpaired supervision. Given a flow prediction at time $t$, we estimate the clean target image as $\hat{x}_0=x_t+(1-t)v_{\theta}(x_t,t \mid s)$. The estimated image is passed through the frozen structure encoder-decoder, and the recovered structural map is compared with the conditioning structure:
\begin{equation}
\mathcal{L}_{\mathrm{struct}}
=
1 -
\mathrm{Dice}
\left(
D_{\phi}(E_{\omega}(\hat{x}_0)),
m_s
\right),
\end{equation}
where $m_s$ denotes the structure map used to define the condition, such as $m_B$ for the target self-domain branch and $\hat{m}_A$ for the source cross-domain branch. This loss encourages the generator to produce target-domain images whose recovered structures remain consistent with the conditioning structure. 

The overall Stage II objective is:
\begin{equation}
\mathcal{L}_{\mathrm{stageII}}
=
\mathcal{L}_{\mathrm{self}}
+
\lambda_{\mathrm{cross}}\mathcal{L}_{\mathrm{cross}}
+
\lambda_{\mathrm{struct}}\mathcal{L}_{\mathrm{struct}}.
\end{equation}
The self-domain branch learns target-domain appearance generation from real target images, the cross-domain branch enables unpaired source-to-target translation through structure-matched soft targets, and the structure consistency loss encourages the generated image to preserve the specified anatomical structure.

%%%%%%%%%%%%%%%%%%%%%%%%%%%%%%%%%%%%%
\input{figures/inference}
%%%%%%%%%%%%%%%%%%%%%%%%%%%%%%%%%%%%%

\subsection{Inference}
\label{sec:traininfer}
% \saum{You will also have to provide steps for (a) Stage I and (b) Stage II too. For eg: do you train $E_\omega$ first with mask images $m$ and then later with $x_A, x_B$, OR, all jointly? Similarly, do you train self-branch and cross-branch separately or jointly?}

We illustrate the inference procedure in Fig.~\ref{fig:inference}. During inference, only the source image $x_A$ is available, and our aim is to generate the corresponding target-domain image $\hat{x}_B$. From $x_A$, we first predict its structural map $\hat{m}_A = D_{\phi}(E_{\omega}(x_A))$. This map is then \textit{re-encoded} into the shared structure space to obtain the flow model condition, $s_A = E_{\omega}(\hat{m}_A)$. Starting from Gaussian noise, the translated target-domain image is generated by solving the conditional flow ODE: $\frac{dx_t}{dt}=v_{\theta}(x_t,t \mid s_A)$ from $t=0$ to $t=1$. 
% \begin{equation}
% \frac{d x_t}{d t}
% =
% v_{\theta}(x_t,t \mid s_A),
% \quad
% t: 0 \rightarrow 1.
% \end{equation}
The final output is $\hat{x}_B = x_{t=1}$. 
No target image/mask or prototype retrieval is required during inference.

%% file: figures/pipeline.tex
\begin{figure*}[!t]
    \centering
    \vspace{-15pt}
    \includegraphics[width=0.95\textwidth]{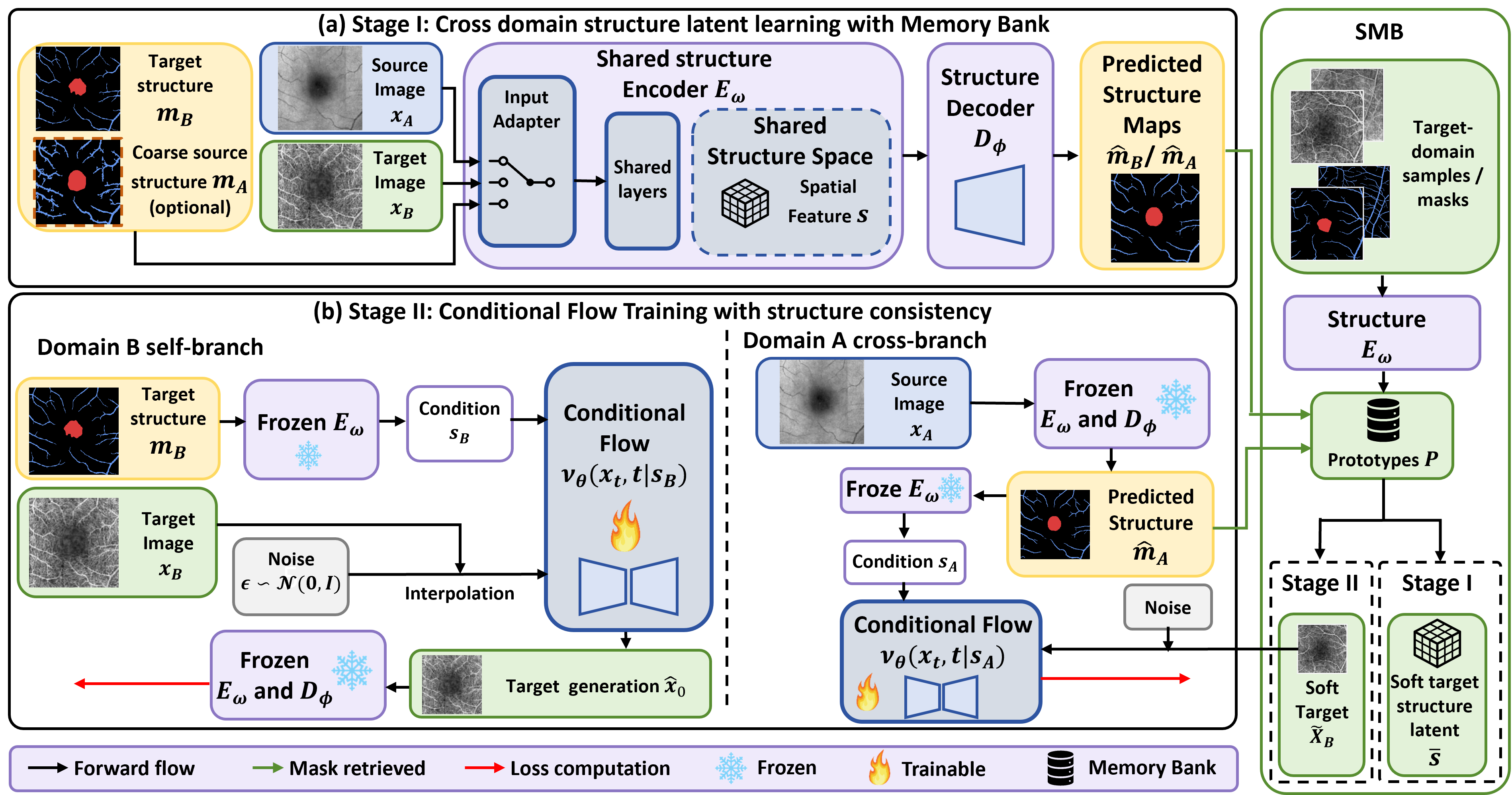}
    \caption{
    Overview of the training procedure of our proposed two-stage framework. 
    Stage I learns a shared structure encoder-decoder that maps images and structural masks into a common structure space, with memory-bank-based soft supervision for source images. 
    After freezing the encoder-decoder, Stage II trains a conditional flow model for target-domain generation. 
    The self-branch learns target-domain appearance rendering, while the cross-branch uses source-derived structures and memory-bank prototypes for cross-domain synthesis.
    }
    \label{fig:pipeline}
    \vspace{-16pt}
\end{figure*}

%% file: figures/inference.tex
\begin{wrapfigure}[8]{r}{0.55\linewidth}
    \vspace{-10pt}
    \centering
    \includegraphics[width=\linewidth]{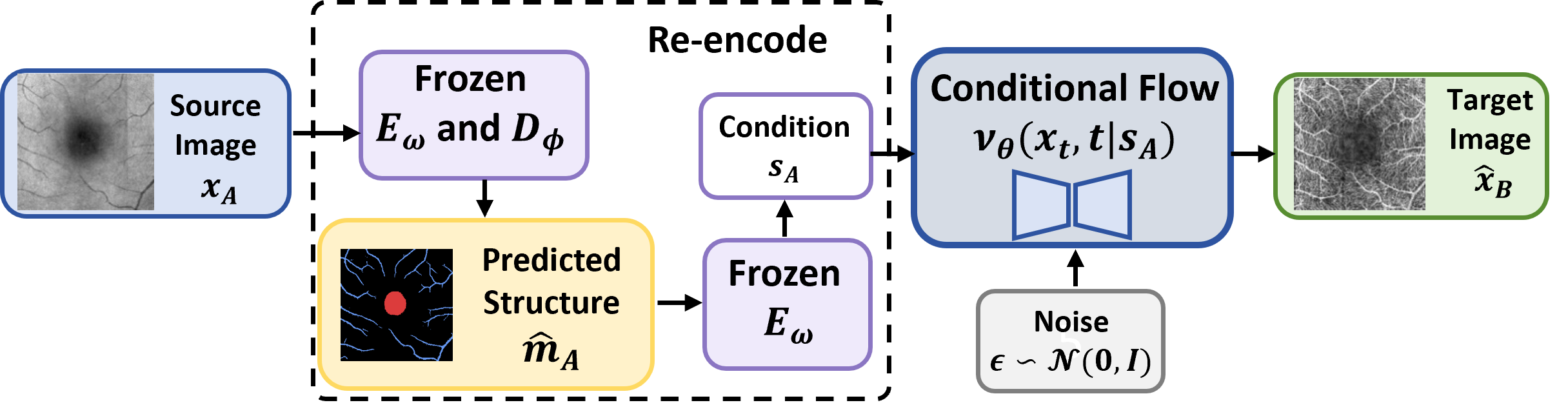}
    \vspace{-17pt}
    \caption{
    Inference pipeline of our structure-conditioned generation framework.}
    \label{fig:inference}
    \vspace{-10pt}
\end{wrapfigure}

%% file: sections/experiment.tex
\section{Experiments}
\label{sec:experiments}

% \subsection{Experimental Setup}

\parag{Datasets.}
We used the \textbf{OCTA-500}~\cite{li2024octa} dataset for 2D retinal OCT-to-OCTA translation. 
The original projections have resolutions of $304 \times 304$ or $400 \times 400$ pixels. 
We split the dataset at the case level into 450 training and 50 test
cases, keeping all projections from each case in the same partition. 
In Stage I, the official full, ILM-OPL, and OPL-BM projections were used as natural augmentation on the same split, yielding 1,350 training and 150 test projection images; in Stage II, only the full projection was used, yielding 450 training and 50 test images. All images were resized to $256 \times 256$. The target-domain mask $m_B$ corresponds to the official OCTA large-vessel and FAZ annotations. We also used the Task 1 brain subset of \textbf{SynthRAD2023}~\cite{thummerer2023synthrad2023} for MRI-to-CT synthesis. The released data includes 180 subjects with registered MR-CT volumes and 30 MR-only validation subjects; since validation CT is unavailable, we randomly selected 20 registered subjects as an internal test set. We extracted 9,646 2D axial slices from the supratentorial brain and skull regions, yielding 8,685 training and 961 test slices. 
All slices were resized with aspect-ratio preservation and padded to $256 \times 256$. 
Anatomical labels were generated using TotalSegmentator~\cite{wasserthal2023totalsegmentator} and merged into six coarse classes. 
Here, the target-domain mask $m_B$ corresponds to CT anatomical labels, while the source-domain coarse mask $m_A$ corresponds to MR coarse anatomical labels. Within each training stage, source and target images for our method
are drawn from non-overlapping cases.
Further details are provided in the Appendix~\ref{sec:sup_data}.

\parag{Baselines.}
We compare our method with representative image-to-image translation baselines under both paired and unpaired settings, covering GAN-based, contrastive, disentanglement-based, diffusion-based, and bridge-based methods. 
For unpaired translation, we include CycleGAN~\cite{zhu2017unpaired}, CUT~\cite{park2020contrastive}, MUNIT~\cite{huang2018multimodal}, SynDiff~\cite{syndiff}, and UNSB~\cite{kim2024unpaired}. 
CycleGAN is a cycle-consistent GAN, CUT uses patch-wise contrastive learning to preserve content correspondence, and MUNIT disentangles content and style representations for multimodal translation. 
In addition, SynDiff introduces an adversarial diffusion framework for unpaired medical image translation, while UNSB models unpaired translation through a Neural Schr\"odinger Bridge formulation. 
For paired translation, we compare against Pix2pix~\cite{isola2017image}, DDBM~\cite{zhou2023denoising}, and MuTri~\cite{chen2025mutri}. 
Pix2pix serves as a classical conditional GAN baseline, DDBM learns a diffusion bridge between paired endpoint distributions, and MuTri~\cite{chen2025mutri} is a recent 3D OCT-to-OCTA translation framework based on VQ-VAE~\cite{van2017neural} and multi-view tri-alignment. We use official implementations and recommended configurations,
with common train/test partitions, preprocessing, resolution, and
evaluation protocols.

\parag{Evaluation Metrics.}
% We use PSNR and SSIM~\cite{hore2010image,wang2004image} to evaluate translated images against the ground-truth target images, reporting metrics over the whole image and task-specific regions. For OCT-to-OCTA translation, region-wise metrics are computed on vessel and FAZ regions; for brain MR-to-CT translation, they are computed on soft-tissue and bone regions, where soft tissue includes the cerebrum, ventricle, thalamus, caudate, and lentiform. All results are reported as mean $\pm$ 95\% confidence interval over the test set, with SSIM shown as a percentage for readability. We also use Dice~\cite{dice1945measures} to evaluate Stage-I structure preservation by comparing the decoded OCT, decoded OCTA, and re-encoded OCT structures with the target-domain mask $m_B$, and measure cross-domain structural consistency by computing Dice between the decoded OCT and OCTA structure masks.
We evaluate translation quality using PSNR and SSIM~\cite{hore2010image,wang2004image} against ground-truth target images, reported over the whole image and task-specific regions: vessel/FAZ for OCT-to-OCTA and soft-tissue/bone for MR-to-CT. Soft tissue includes the cerebrum, ventricle, thalamus, caudate, and lentiform. Results are reported as mean $\pm$ 95\% confidence interval, with SSIM shown as a percentage. We also use Dice~\cite{dice1945measures} to assess Stage-I consistency between decoded/re-encoded structure masks and the target-domain mask $m_B$.

% \saum{mention the Table 3 DICE eval pipeline too. Since the claim of the paper is structure-preserving, this is the most important evaluation. Can also try including 1-betti for retinal vessel, and 0/1-betti for brain}

\subsection{Results}
\label{sec:results}

% \saum{We may need to quantitatively evaluate intermediate outputs in your pipeline....similar to how SPIGen reviewers asked to evaluate pseudo mask}

% \saum{Sensitivity analysis of different $\lambda$ values}

% \saum{Mention inference time of your method in seconds and other methods for a particular image size}

%%%%%%%%%%%%%%%Table 1%%%%%%%%%%%%%%%%%%%%
\input{tables/main-result}
%%%%%%%%%%%%%%%Table 1%%%%%%%%%%%%%%%%%%%%

%%%%%%%%%%%%%%%Table 2%%%%%%%%%%%%%%%%%%%%
\input{tables/main-result2}
%%%%%%%%%%%%%%%Table 2%%%%%%%%%%%%%%%%%%%%

% %%%%%%%%%%%%%%%Table 5%%%%%%%%%%%%%%%%%%%%
% \input{tables/bank_K}
% %%%%%%%%%%%%%%%Table 5%%%%%%%%%%%%%%%%%%%

% %%%%%%%%%%%%%%%Table 6%%%%%%%%%%%%%%%%%%%%
% \input{tables/loss_cycle}
% %%%%%%%%%%%%%%%Table 6%%%%%%%%%%%%%%%%%%%

% %%%%%%%%%%%%%%%Table 7%%%%%%%%%%%%%%%%%%%%
% \input{tables/loss_struct}
% %%%%%%%%%%%%%%%Table 6%%%%%%%%%%%%%%%%%%%

%%%%%%%%%%%%%%%Table 4%%%%%%%%%%%%%%%%%%%%
\input{tables/stageI_analysis}
%%%%%%%%%%%%%%%Table 4%%%%%%%%%%%%%%%%%%%
\parag{Quantitative results.} 
Although our method is trained in an unpaired setting, evaluation is conducted on test pairs by comparing each translated image with its ground-truth target image using PSNR and SSIM over the whole image and task-specific regions. 
Tab.~\ref{tab:translation_results} shows the results on OCT-to-OCTA translation. 
Our method consistently outperforms existing unpaired baselines on the whole image, vessel region, and FAZ region, demonstrating improved vascular recovery and structural correspondence. Notably, despite using no paired source-target supervision during training, it also surpasses paired baselines, improving over the best paired result by up to $11.2\%$ in PSNR and $26.4\%$ in SSIM. Tab.~\ref{tab:brain_translation_results} further shows consistent gains on brain MR-to-CT translation. Compared with the best unpaired baseline, our method improves PSNR by $16.5\%$, $19.7\%$, and $23.1\%$ on the whole image, soft-tissue region, and bone region, respectively, with SSIM gains across all regions. 
It also outperforms paired baselines on all metrics, achieving PSNR gains of $3.2\%$--$4.6\%$ with consistent SSIM improvements. 
Overall, these results show that our structure-conditioned formulation improves both target-domain fidelity and structure preservation, especially in anatomy-critical regions.

%%%%%%%%%%%%%%%Figure 4 9%%%%%%%%%%%%%%%%%%%%
\input{figures/main_res}
%%%%%%%%%%%%%%%Figure 4  9%%%%%%%%%%%%%%%%%%%
\parag{Qualitative Results.}
Fig.~\ref{fig:qualitative_comparison} presents qualitative comparisons under both unpaired and paired settings on OCT-to-OCTA and brain MR-to-CT translation. For each method, the first row shows the translated target-domain image, and the second row shows the MAE map between the prediction and ground-truth target image. Overall, our method produces more realistic target-domain images with better anatomical preservation and lower reconstruction errors. 
As highlighted by the red boxes, existing baselines often lose fine structures, such as missing vessel branches in OCTA images and inconsistent ventricle shapes in brain CT images; these errors are also reflected by stronger MAE responses around vessel regions and anatomical boundaries. In contrast, our method better preserves these structures, generating more continuous vascular patterns for OCT-to-OCTA translation and clearer CT-like soft-tissue and bone structures for brain MR-to-CT translation. These improvements come from explicitly separating structure preservation from appearance generation: the shared structure encoder provides domain-invariant anatomical guidance, while the target-domain flow model renders realistic appearance under this guidance. 
More qualitative results are provided in the Appendix~\ref{sec:qualitative_supp}.

\parag{Ablation study.} 
%%%%%%%%%%%%%%%Figure 5 9%%%%%%%%%%%%%%%%%%%%
\input{figures/ablation_parameter}
%%%%%%%%%%%%%%%Figure 5  9%%%%%%%%%%%%%%%%%%%

%%%%%%%%%%%%%%%Table 5%%%%%%%%%%%%%%%%%%%%
\input{tables/cross-center}
%%%%%%%%%%%%%%%Table 5%%%%%%%%%%%%%%%%%%%

%%%%%%%%%%%%%%%Table 6%%%%%%%%%%%%%%%%%%%%
\input{tables/downstream-class}
%%%%%%%%%%%%%%%Table 6%%%%%%%%%%%%%%%%%%%

%%%%%%%%%%%%%%%Table 6%%%%%%%%%%%%%%%%%%%%
\input{tables/independent-structure}
%%%%%%%%%%%%%%%Table 6%%%%%%%%%%%%%%%%%%%
We analyze the effects of $\lambda_{\mathrm{cycle}}$, $\lambda_{\mathrm{struct}}$, and the number of retrieved prototypes $K$ on OCT-to-OCTA translation, as shown in Fig.~\ref{fig:ablation_line_charts}. Removing $\mathcal{L}_{\mathrm{cycle}}$ clearly degrades performance, while non-zero values yield stable results, indicating that latent cycle consistency helps suppress appearance cues and align OCT- and OCTA-derived structure conditions. 
For $\lambda_{\mathrm{struct}}$, a moderate weight improves structural guidance, but an overly large value reduces performance, suggesting that excessive regularization can bias the conditional flow away from the target OCTA distribution. For the structure bank, a single prototype provides an unreliable hard target, whereas multiple prototypes offer more robust soft structural guidance and stable performance across $K$. 
Overall, the results show that our framework benefits from cycle consistency, a non-excessive structure constraint, and multi-prototype aggregation, while remaining relatively insensitive to exact hyperparameter settings.

Tab.~\ref{tab:structure_dice_ablation} evaluates the Stage I shared structure encoder. 
Let 
$\hat{m}_A=D_{\phi}(E_{\omega}(x_A))$, 
$\hat{m}_B=D_{\phi}(E_{\omega}(x_B))$, and 
$\widetilde{m}_A=D_{\phi}(E_{\omega}(\hat{m}_A))$. 
We define 
$\mathrm{Dice}_{\mathrm{OCT}}=\mathrm{Dice}(\hat{m}_A,m_B)$, 
$\mathrm{Dice}_{\mathrm{OCTA}}=\mathrm{Dice}(\hat{m}_B,m_B)$, 
$\mathrm{Dice}_{\mathrm{re}}=\mathrm{Dice}(\widetilde{m}_A,m_B)$, and 
$\mathrm{Dice}_{\mathrm{pair}}=\mathrm{Dice}(\hat{m}_A,\hat{m}_B)$.  
% \saum{i think you should use math variables instead of English here. for eg: $\mathrm{Dice}_{\mathrm{OCTA}} = \text{Dice}(D(E(x_A)), m_B)$ etc. It is totally not clear in English what Diceoct Diceocta Dicepair Dicere etc mean, just use math} 
Removing the source coarse mask input $m_A$ causes only limited changes: although the full model performs better on $\mathrm{Dice}_{\mathrm{OCT}}$, $\mathrm{Dice}_{\mathrm{re}}$, and $\mathrm{Dice}_{\mathrm{pair}}$, the variant without $m_A$ remains comparable and even improves $\mathrm{Dice}_{\mathrm{OCTA}}$. 
This indicates that $m_A$ provides auxiliary source-side guidance but is not the dominant factor for structure extraction. 
In contrast, removing $\mathcal{L}_{\mathrm{cycle}}$ more clearly reduces cross-domain structural consistency, with $\mathrm{Dice}_{\mathrm{pair}}$ dropping from 72.62 to 69.56. 
Overall, the encoder can learn reasonable structural representations even without $m_A$, while cycle consistency is more important for aligning OCT and OCTA structures in the shared space. Qualitative mask visualizations are provided in the Appendix Fig.~\ref{fig:stage1_structure_qualitative}.

\parag{Cross-center generalization.}
We retrain the evaluated methods on SynthRAD2023 centers A and B
and test on unseen center C without adaptation. No data or prototypes
from center C are used during training or model selection.
As shown in Tab.~\ref{tab:center_heldout}, our method achieves
the best mean performance across all three anatomical regions.

% %%%%%%%%%%%%%%%Table 5%%%%%%%%%%%%%%%%%%%%
% \input{tables/cross-center}
% %%%%%%%%%%%%%%%Table 5%%%%%%%%%%%%%%%%%%%

% %%%%%%%%%%%%%%%Table 6%%%%%%%%%%%%%%%%%%%%
% \input{tables/downstream-class}
% %%%%%%%%%%%%%%%Table 6%%%%%%%%%%%%%%%%%%%

% %%%%%%%%%%%%%%%Table 6%%%%%%%%%%%%%%%%%%%%
% \input{tables/independent-structure}
% %%%%%%%%%%%%%%%Table 6%%%%%%%%%%%%%%%%%%%

\parag{Downstream and structural evaluation.}
We evaluate generated OCTA using a normal-versus-abnormal classifier
trained exclusively on acquired OCTA from 450 training cases and
frozen for testing on 50 held-out cases.
As shown in Tab.~\ref{tab:diagnostic_evaluation}, our method achieves
the highest balanced accuracy among translated inputs and the lowest
probability deviation and decision-flip rate relative to acquired OCTA.
On the same cases, we assess vessel fidelity using a frozen evaluator
independent of translation training and model selection.
Tab.~\ref{tab:independent_vessel_evaluation} shows that our method
achieves the highest mean vessel Dice and lowest mean structural
errors, with comparable clDice~\cite{shit2021cldice} to MuTri.

%% file: tables/main-result.tex
\begin{table}[t]
\vspace{-12pt}
\centering
\caption{
Comparison on the OCT to OCTA translation task. We report mean $\pm$ 95\% confidence interval for whole-image, vessel-region, and FAZ-region metrics. \textbf{Bold} denotes the best result, significantly outperforming the best unpaired baseline.
%$\checkmark$ indicates paired training.
}
\label{tab:translation_results}
\scriptsize
\setlength{\tabcolsep}{2.5pt}
\renewcommand{\arraystretch}{1.08}
\resizebox{0.8\linewidth}{!}{%
\begin{tabular}{@{}lccccccc@{}}
\toprule
\multirow{2}{*}{\textbf{Method}} & 
\multirow{2}{*}{\textbf{Paired}} &
\multicolumn{2}{c}{\textbf{Whole Image}} &
\multicolumn{2}{c}{\textbf{Vessel}} &
\multicolumn{2}{c}{\textbf{FAZ}} \\
\cmidrule(lr){3-4}
\cmidrule(lr){5-6}
\cmidrule(lr){7-8}
& &
\textbf{PSNR} $\uparrow$ & \textbf{SSIM} $\uparrow$ &
\textbf{PSNR} $\uparrow$ & \textbf{SSIM} $\uparrow$ &
\textbf{PSNR} $\uparrow$ & \textbf{SSIM} $\uparrow$ \\
\midrule
CycleGAN~\cite{zhu2017unpaired} & $\times$ & $14.21{\pm}0.18$ & $5.71{\pm}1.11$ & $9.74{\pm}0.27$ & $6.34{\pm}1.81$ & $16.79{\pm}0.77$ & $23.15{\pm}1.87$ \\
CUT~\cite{park2020contrastive} & $\times$ & $13.62{\pm}0.22$ & $7.91{\pm}2.01$ & $14.17{\pm}0.37$ & $8.24{\pm}1.47$ & $15.79{\pm}0.54$ & $19.87{\pm}2.23$ \\
MUNIT~\cite{huang2018multimodal} & $\times$ & $16.12{\pm}0.16$ & $15.44{\pm}1.02$ & $13.81{\pm}0.35$ & $32.82{\pm}2.41$ & $16.51{\pm}0.75$ & $10.57{\pm}3.08$ \\
SynDiff~\cite{syndiff} & $\times$ & $14.41{\pm}0.20$ & $6.48{\pm}1.46$ & $12.06{\pm}0.30$ & $5.70{\pm}1.52$ & $18.09{\pm}0.81$ & $27.99{\pm}2.02$ \\
UNSB~\cite{kim2024unpaired} & $\times$ & $14.31{\pm}0.23$ & $7.16{\pm}1.31$ & $10.19{\pm}0.36$ & $5.32{\pm}1.62$ & $18.27{\pm}0.88$ & $32.46{\pm}2.42$ \\
\midrule
DDBM~\cite{zhou2023denoising} & $\checkmark$ & $15.84{\pm}0.21$ & $27.15{\pm}0.73$ & $14.84{\pm}0.42$ & $52.46{\pm}3.27$ & $20.18{\pm}1.15$ & $33.01{\pm}2.55$ \\
Pix2pix~\cite{isola2017image} & $\checkmark$ & $15.61{\pm}0.20$ & $23.83{\pm}0.91$ & $14.82{\pm}0.31$ & $50.66{\pm}3.09$ & $20.15{\pm}0.74$ & $29.79{\pm}1.86$ \\
MuTri~\cite{chen2025mutri} & $\checkmark$ & $18.59{\pm}0.46$ & $51.16{\pm}1.76$ & $17.02{\pm}0.49$ & $70.09{\pm}2.73$ & $20.21{\pm}1.02$ & $34.16{\pm}2.88$ \\
\midrule
Ours & $\times$ & $\textbf{19.05}{\pm}\textbf{0.72}$ & $\textbf{64.65}{\pm}\textbf{5.40}$ & $\textbf{18.92}{\pm}\textbf{0.88}$ & $\textbf{75.72}{\pm}\textbf{5.81}$ & $\textbf{20.48}{\pm}\textbf{1.00}$ & $\textbf{35.05}{\pm}\textbf{4.03}$ \\
\bottomrule
\end{tabular}%
}
\vspace{-8pt}
\end{table}

%% file: tables/main-result2.tex
\begin{table}[t]
\vspace{-8pt}
\centering
\caption{
 Comparison on the brain MR-to-CT translation task. 
We report mean $\pm$ 95\% confidence interval for whole-image, soft-tissue-region, and bone-region metrics. \textbf{Bold} denotes the best result, significantly outperforming the best unpaired baseline.
% $\checkmark$ indicates paired training.
}
\label{tab:brain_translation_results}
\scriptsize
\setlength{\tabcolsep}{2.5pt}
\renewcommand{\arraystretch}{1.08}
\resizebox{0.8\linewidth}{!}{%
\begin{tabular}{@{}lccccccc@{}}
\toprule
\multirow{2}{*}{\textbf{Method}} & 
\multirow{2}{*}{\textbf{Paired}} &
\multicolumn{2}{c}{\textbf{Whole Image}} &
\multicolumn{2}{c}{\textbf{Soft Tissue}} &
\multicolumn{2}{c}{\textbf{Bone}} \\
\cmidrule(lr){3-4}
\cmidrule(lr){5-6}
\cmidrule(lr){7-8}
& &
\textbf{PSNR} $\uparrow$ & \textbf{SSIM} $\uparrow$ &
\textbf{PSNR} $\uparrow$ & \textbf{SSIM} $\uparrow$ &
\textbf{PSNR} $\uparrow$ & \textbf{SSIM} $\uparrow$ \\
\midrule
CycleGAN~\cite{zhu2017unpaired} & $\times$      & $19.48{\pm}0.10$ & $88.52{\pm}0.16$ & $23.76{\pm}0.18$ & $60.56{\pm}0.62$ & $9.86{\pm}0.09$ & $22.60{\pm}1.05$ \\
CUT~\cite{park2020contrastive}      & $\times$      & $20.52{\pm}0.11$ & $90.13{\pm}0.13$& $20.95{\pm}0.18$ & $59.25{\pm}0.57$ & $9.81{\pm}0.09$ & $22.04{\pm}0.90$ \\
MUNIT~\cite{huang2018multimodal}     & $\times$      & $21.79{\pm}0.13$ & $90.80{\pm}0.16$ & $21.78{\pm}0.18$ & $51.02{\pm}0.61$ & $11.75{\pm}0.09$ & $44.14{\pm}1.04$ \\
SynDiff~\cite{syndiff}  & $\times$      & $24.58{\pm}0.14$ & $94.50{\pm}0.12$ & $23.33{\pm}0.19$ & $54.11{\pm}0.86$ & $14.31{\pm}0.12$ & $66.72{\pm}0.83$ \\
UNSB~\cite{kim2024unpaired}     & $\times$      & $19.51{\pm}0.10$ & $88.34{\pm}0.17$ & $20.90{\pm}0.21$ & $61.02{\pm}0.51$ & $8.89{\pm}0.09$ & $11.70{\pm}1.15$ \\
\midrule
DDBM~\cite{zhou2023denoising}       & $\checkmark$ & $27.67{\pm}0.14$ & $97.01{\pm}0.09$ & $27.20{\pm}0.12$ & $68.94{\pm}0.52$ & $16.92{\pm}0.13$ & $79.05{\pm}0.60$ \\
Pix2pix~\cite{isola2017image}    & $\checkmark$ & $27.75{\pm}0.14$ & $96.66{\pm}0.15$ & $26.96{\pm}0.13$ & $66.80{\pm}0.49$ & $16.91{\pm}0.13$ & $78.42{\pm}0.65$ \\
\midrule
Ours & $\times$    & $\textbf{28.64}{\pm}\textbf{0.20}$ & $\textbf{97.15}{\pm}\textbf{0.12}$ & $\textbf{28.45}{\pm}\textbf{0.15}$ & $\textbf{69.20}{\pm}\textbf{0.55}$ & $\textbf{17.62}{\pm}\textbf{0.17}$ & $\textbf{80.30}{\pm}\textbf{0.89}$ \\
\bottomrule
\end{tabular}%
}
\vspace{-18pt}
\end{table}

%% file: tables/stageI_analysis.tex
\begin{wraptable}[6]{r}{0.45\linewidth}
    \vspace{-10pt}
    \centering
    \scriptsize
    \setlength{\tabcolsep}{3.5pt}
    \renewcommand{\arraystretch}{1.12}
    \caption{
    Stage I effectiveness of the shared structure encoder.
    }
    \vspace{-8pt}
    \label{tab:structure_dice_ablation}
    \resizebox{\linewidth}{!}{
    \begin{tabular}{@{}lcccc@{}}
    \toprule
    \textbf{Setting} 
    & \textbf{$\mathrm{Dice}_{\mathrm{OCT}}$}$\uparrow$
    & \textbf{$\mathrm{Dice}_{\mathrm{OCTA}}$}$\uparrow$
    & \textbf{$\mathrm{Dice}_{\mathrm{re}}$}$\uparrow$
    & \textbf{$\mathrm{Dice}_{\mathrm{pair}}$}$\uparrow$ \\
    \midrule
    w/o $m_A$ 
    & $73.33{\pm}0.02$ & $75.78{\pm}0.03$ & $75.10{\pm}0.04$ & $71.23{\pm}0.03$ \\
    w/o $\mathcal{L}_{\mathrm{cycle}}$ 
    & $72.96{\pm}0.04$ & $74.66{\pm}0.03$ & $75.06{\pm}0.04$ & $69.56{\pm}0.03$ \\
    Ours 
    & $\textbf{75.89}{\pm}\textbf{0.03}$ 
    & $\textbf{74.75}{\pm}\textbf{0.02}$ 
    & $\textbf{76.92}{\pm}\textbf{0.04}$ 
    & $\textbf{72.62}{\pm}\textbf{0.05}$ \\
    \bottomrule
    \end{tabular}
    }
    \vspace{-20pt}
\end{wraptable}

%% file: figures/main_res.tex
\begin{figure}[t]
    \centering
    \includegraphics[width=0.9\linewidth]{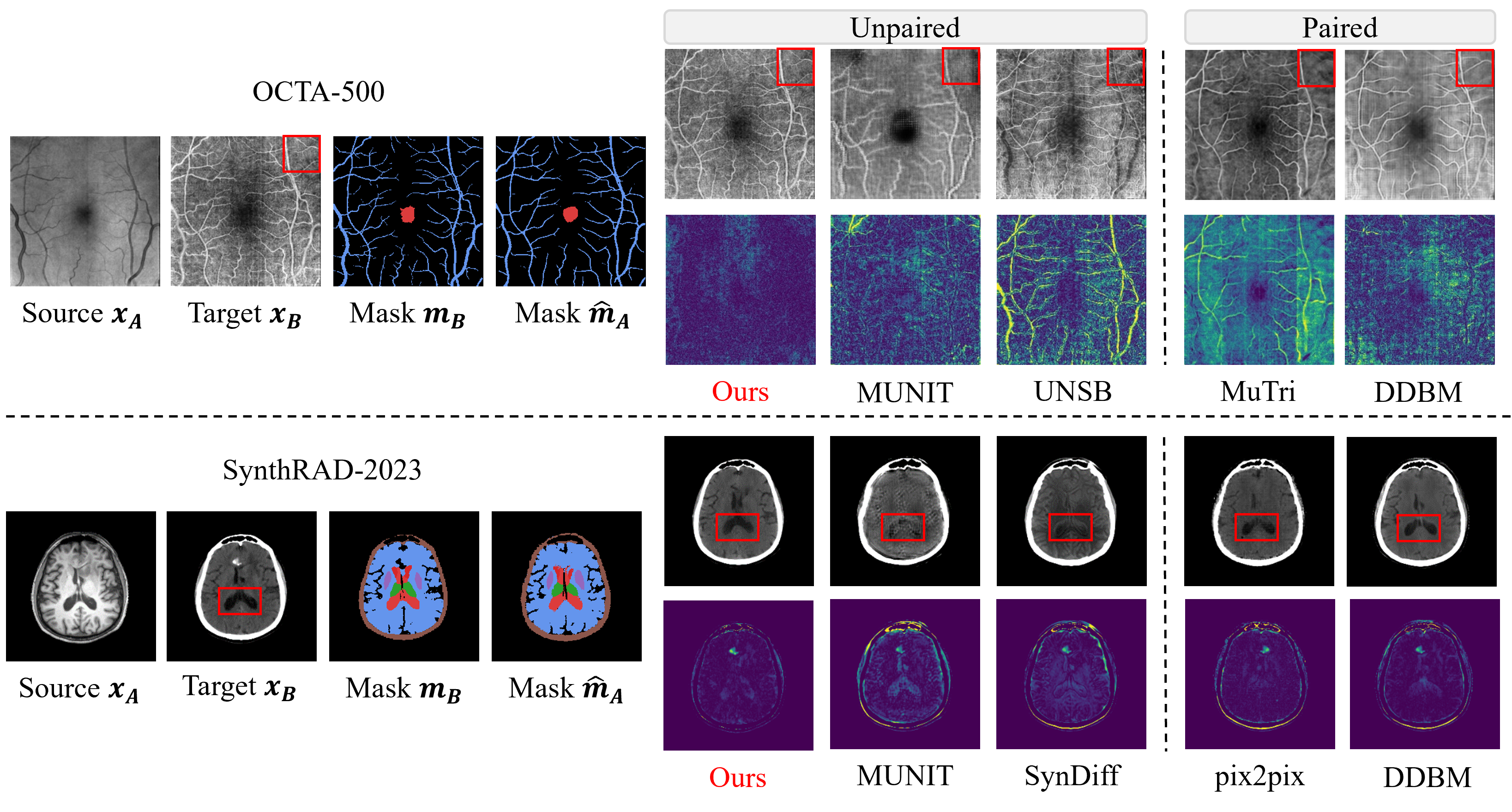}
    \vspace{-12pt}
    \caption{
    \textbf{Qualitative comparison.} 
    For each method, the first row shows the translated target-domain image and the second row shows the mean absolute error (MAE) map against the ground truth. 
    For our method, the source structural condition is obtained as $\hat{m}_A = D_{\phi}(E_{\omega}(x_A))$. 
    For visualization, CT images are re-windowed to a brain window of $[0,100]$ HU to enhance soft-tissue contrast, while quantitative metrics are computed in the original normalized CT space. Our method better preserves anatomy and reduces errors in structure-critical regions (red boxes).
    }
    \label{fig:qualitative_comparison}
    \vspace{-8pt}
\end{figure}

%% file: figures/ablation_parameter.tex
\begin{figure}[!t]
    \centering
    \captionsetup[subfigure]{justification=centering}
    \setlength{\abovecaptionskip}{3pt}
    \setlength{\belowcaptionskip}{-2pt}

    \begin{subfigure}[t]{0.32\linewidth}
        \centering
        \includegraphics[
            width=\linewidth,
            trim=8 3mm 8 8,
            clip
        ]{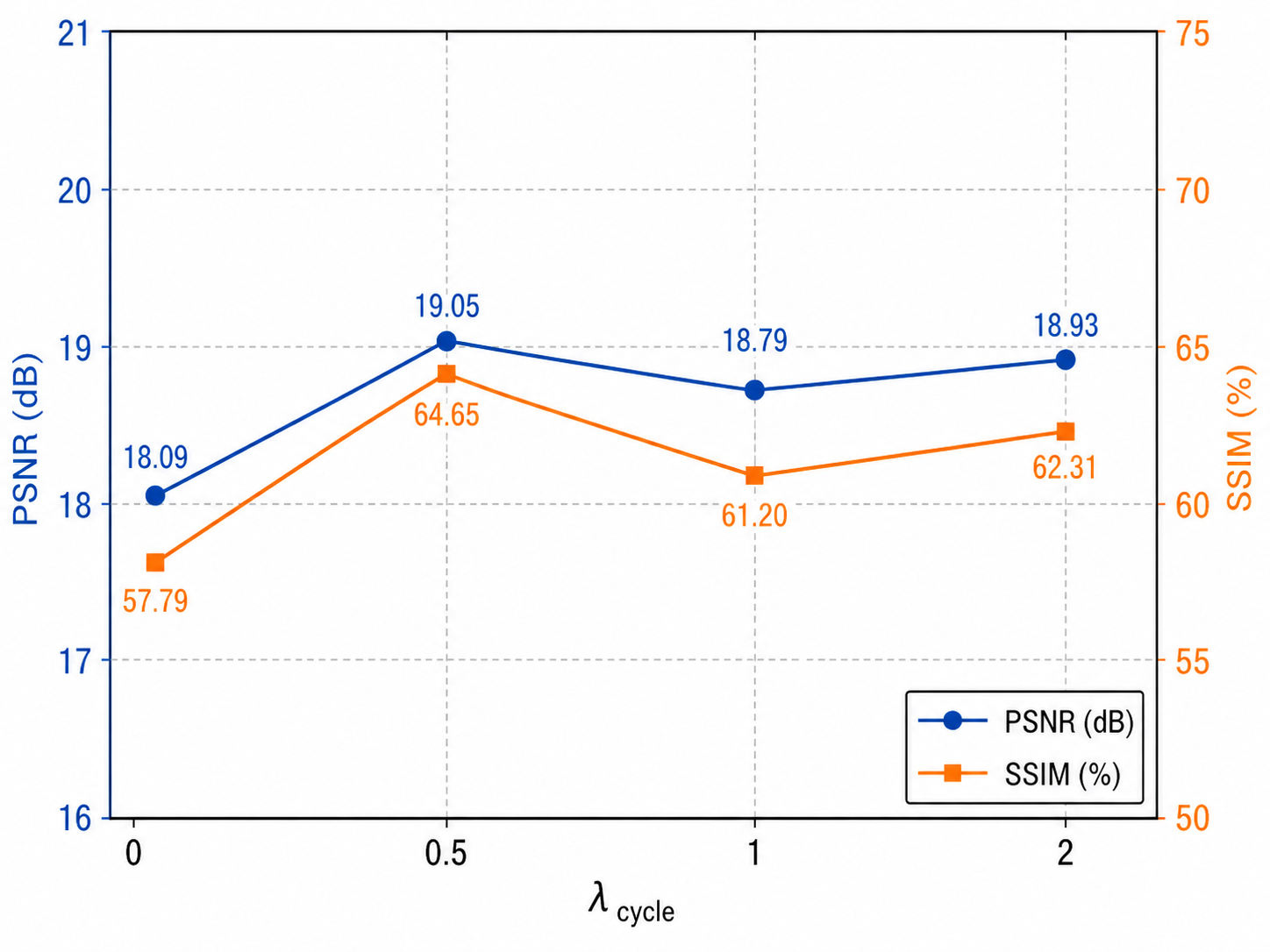}
        \caption{$\lambda_{\mathrm{cycle}}$}
        \label{fig:ablation_lambda_cycle}
    \end{subfigure}
    \hfill
    \begin{subfigure}[t]{0.32\linewidth}
        \centering
        \includegraphics[
            width=\linewidth,
            trim=8 3mm 8 8,
            clip
        ]{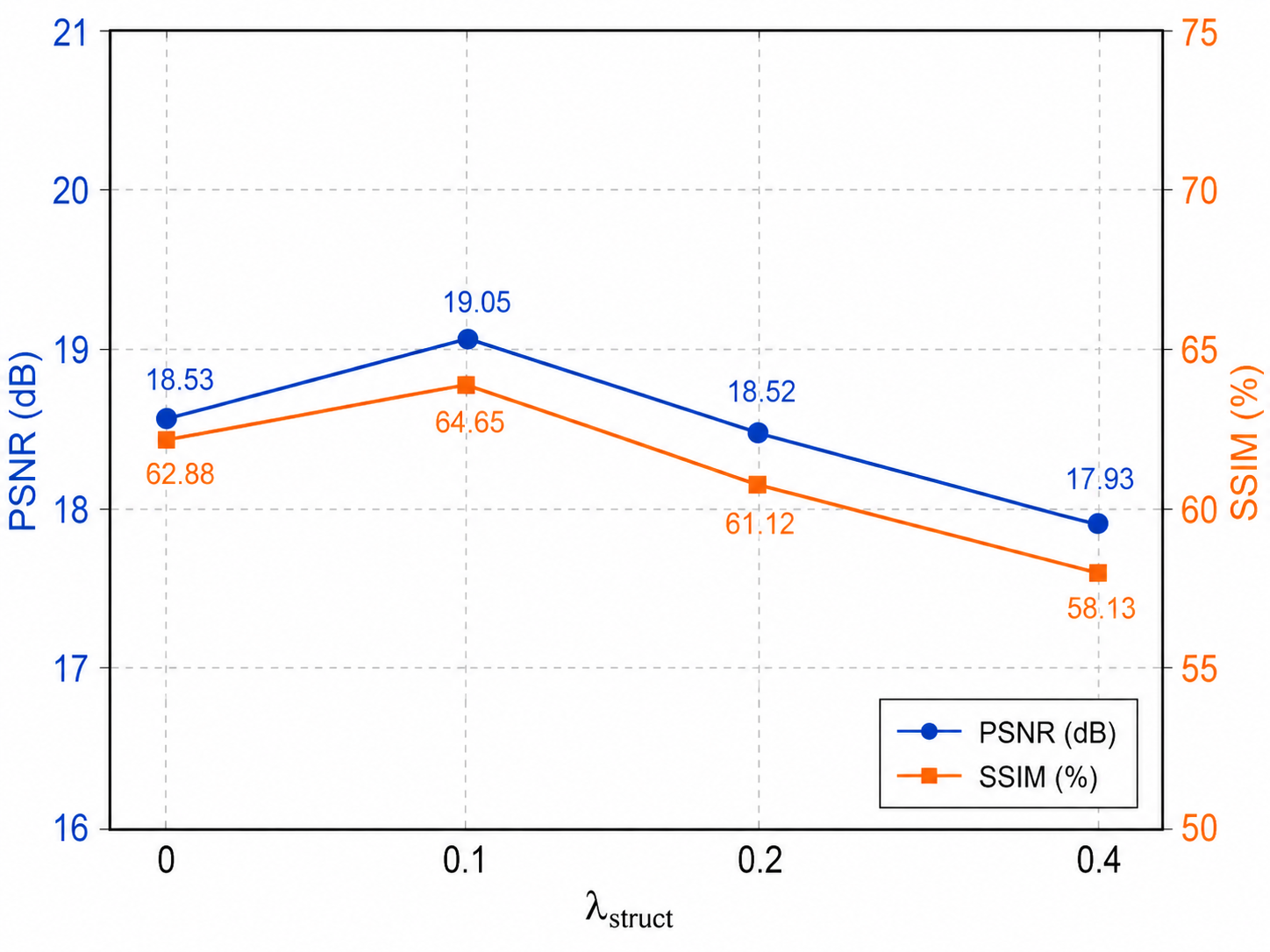}
        \caption{$\lambda_{\mathrm{struct}}$}
        \label{fig:ablation_lambda_struct}
    \end{subfigure}
    \hfill
    \begin{subfigure}[t]{0.32\linewidth}
        \centering
        \includegraphics[
            width=\linewidth,
            trim=8 3mm 8 8,
            clip
        ]{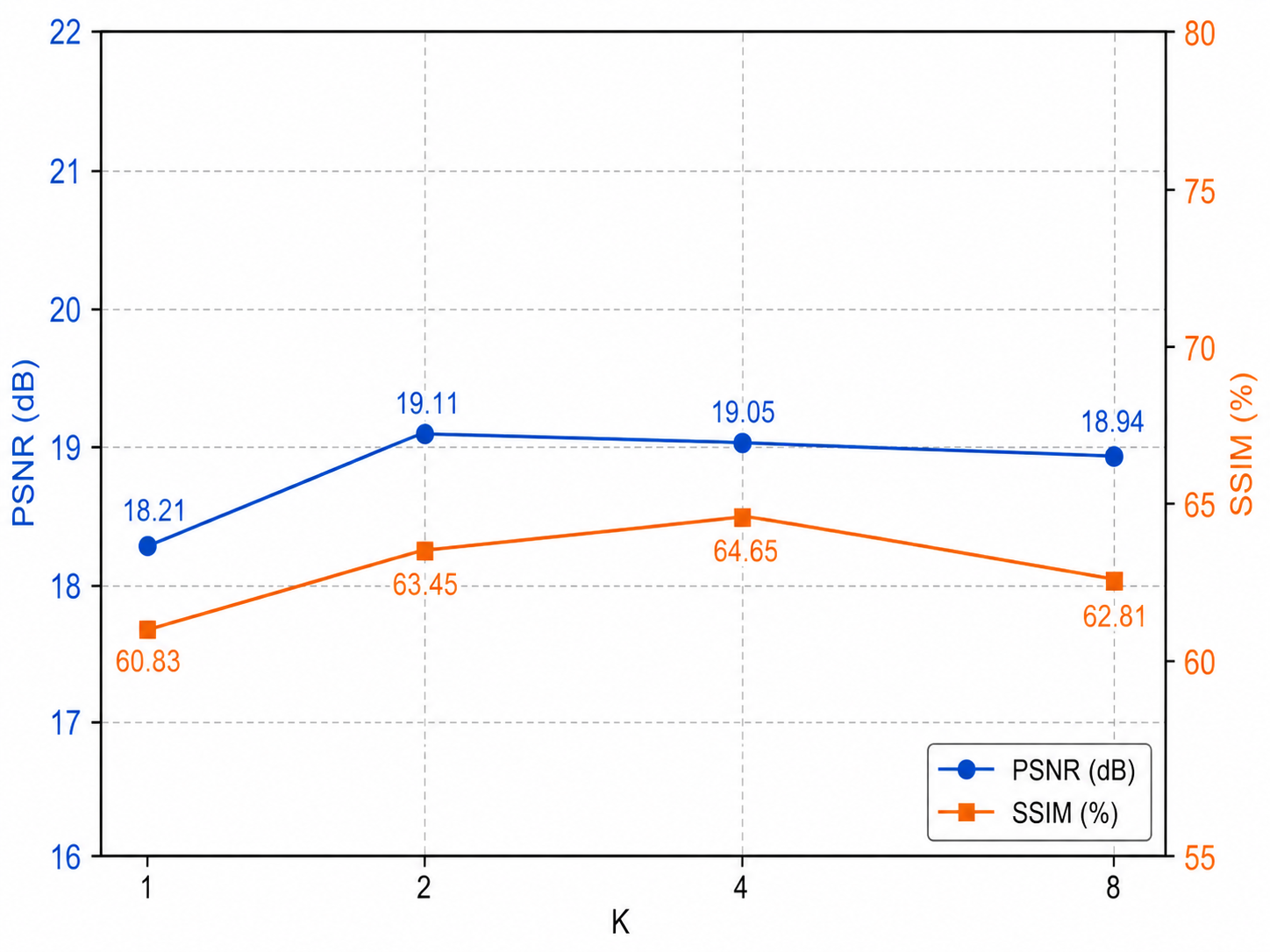}
        \caption{Top-$K$ prototypes}
        \label{fig:ablation_topk_bank}
    \end{subfigure}
    \caption{
    \textbf{Ablation studies on the OCT-to-OCTA translation task.}
    We evaluate the effects of key hyperparameters on translation performance using PSNR and SSIM.
    \textbf{(a)} Varying the weight $\lambda_{\mathrm{cycle}}$ of the cycle loss $\mathcal{L}_{\mathrm{cycle}}$.
    \textbf{(b)} Varying the weight $\lambda_{\mathrm{struct}}$ of the structure consistency loss $\mathcal{L}_{\mathrm{struct}}$ in Stage II.
    \textbf{(c)} Varying the number of retrieved prototypes $K$ in the target-domain structure bank SMB.
    }
    \label{fig:ablation_line_charts}
    \vspace{-16pt}
\end{figure}

%% file: tables/cross-center.tex
\begin{table}
\vspace{-8pt}
\centering
\caption{
Center-held-out MR-to-CT evaluation.
Results are patient-level means; PSNR is in dB and SSIM in \%.
``Paired'' denotes image-pair supervision.
Our method uses target-domain structural masks.
}
\label{tab:center_heldout}
\small
\setlength{\tabcolsep}{4pt}
\begin{tabular}{@{}lccccccc@{}}
\toprule
\textbf{Method} &
\textbf{Paired} &
\multicolumn{2}{c}{\textbf{Whole Image}} &
\multicolumn{2}{c}{\textbf{Soft Tissue}} &
\multicolumn{2}{c}{\textbf{Bone}} \\
\cmidrule(lr){3-4}
\cmidrule(lr){5-6}
\cmidrule(lr){7-8}
& &
PSNR $\uparrow$ & SSIM $\uparrow$ &
PSNR $\uparrow$ & SSIM $\uparrow$ &
PSNR $\uparrow$ & SSIM $\uparrow$ \\
\midrule
SynDiff & No
& 24.10 & 94.68 & 23.18 & 50.71 & 13.71 & 62.31 \\
DDBM & Yes
& 25.13 & 95.79 & 23.27 & 58.71 & 14.10 & 67.49 \\
Ours & No
& \textbf{27.98} & \textbf{97.05}
& \textbf{27.15} & \textbf{59.17}
& \textbf{17.14} & \textbf{80.45} \\
\bottomrule
\end{tabular}
\vspace{-6pt}
\end{table}

%% file: tables/downstream-class.tex
\begin{table}
\vspace{-6pt}
\centering
\caption{
Downstream classification on OCTA-500.
Brackets denote 95\% case-level bootstrap confidence intervals.
$|\Delta p|$ and flip rate measure probability and decision changes
relative to acquired OCTA. Bold indicates the best mean among
translated inputs.
}
\label{tab:diagnostic_evaluation}
\small
\setlength{\tabcolsep}{4pt}
\begin{tabular}{@{}lcccc@{}}
\toprule
\textbf{Input} &
\textbf{AUROC} $\uparrow$ &
\textbf{Bal. Acc.} $\uparrow$ &
$\boldsymbol{|\Delta p|}$ $\downarrow$ &
\textbf{Flip rate} $\downarrow$ \\
\midrule
Acquired OCTA
& $0.928\,[0.848,\,0.984]$
& $0.850\,[0.750,\,0.944]$
& -- & -- \\
\midrule
MuTri
& $0.702\,[0.555,\,0.840]$
& $0.533\,[0.469,\,0.612]$
& $0.288$ & $0.30$ \\
SynDiff
& $\mathbf{0.888}\,[0.773,\,0.976]$
& $0.533\,[0.468,\,0.612]$
& $0.315$ & $0.30$ \\
UNSB
& $0.525\,[0.353,\,0.700]$
& $0.550\,[0.500,\,0.625]$
& $0.355$ & $0.28$ \\
MUNIT
& $0.455\,[0.297,\,0.619]$
& $0.500\,[0.500,\,0.500]$
& $0.416$ & $0.28$ \\
Ours
& $0.887\,[0.740,\,0.995]$
& $\mathbf{0.867}\,[0.762,\,0.958]$
& $\mathbf{0.132}$ & $\mathbf{0.08}$ \\
\bottomrule
\end{tabular}
\vspace{-12pt}
\end{table}

%% file: tables/independent-structure.tex
\begin{table}

\centering
\caption{
Independent vessel evaluation on 50 held-out OCT-to-OCTA cases.
Values are mean $\pm$ 95\% case-level bootstrap CI half-width.
CC denotes connected components. Bold indicates the best mean.
}
\label{tab:independent_vessel_evaluation}
\small
\setlength{\tabcolsep}{3pt}
\renewcommand{\arraystretch}{1.05}
\resizebox{\linewidth}{!}{%
\begin{tabular}{@{}lcccccc@{}}
\toprule
\textbf{Method} &
\textbf{Vessel Dice} $\uparrow$ &
\textbf{clDice} $\uparrow$ &
\textbf{Density Err.} $\downarrow$ &
\textbf{Halluc.} $\downarrow$ &
\textbf{Omiss.} $\downarrow$ &
\textbf{\#CC Err.} $\downarrow$ \\
\midrule
MUNIT
& $0.501\pm0.029$
& $0.493\pm0.032$
& $0.0304\pm0.0037$
& $0.362\pm0.030$
& $0.584\pm0.029$
& $28.9\pm5.4$ \\
MuTri
& $0.787\pm0.017$
& $\mathbf{0.818}\pm0.019$
& $0.0118\pm0.0026$
& $0.162\pm0.010$
& $0.253\pm0.028$
& $25.3\pm7.3$ \\
Ours
& $\mathbf{0.797}\pm0.026$
& $\mathbf{0.818}\pm0.027$
& $\mathbf{0.0113}\pm0.0031$
& $\mathbf{0.152}\pm0.014$
& $\mathbf{0.243}\pm0.036$
& $\mathbf{17.8}\pm6.0$ \\
\bottomrule
\end{tabular}%
}
\vspace{-6pt}
\end{table}

%% file: sections/conclusion.tex
\section{Conclusion}
We proposed a two-stage unpaired medical image translation framework that separates structure representation learning from target-domain generation. 
Stage I learns a shared structure space through mask reconstruction and prototype-based soft supervision to capture modality-agnostic anatomy. 
Stage II freezes the structure encoder-decoder and uses its structural representations to condition a target-domain flow model, enabling appearance synthesis under fixed anatomical guidance. 
Experiments on OCT-to-OCTA and brain MR-to-CT translation show that our method outperforms both unpaired and paired baselines, preserving clinically meaningful structures while generating realistic target-domain images. 
These results highlight the benefit of decoupling structural guidance from appearance rendering for robust unpaired medical image translation.

% \textbf{Acknowledgments.} This research was partially supported by grants NSF CCF-2144901, NIH R01NS143143 and R01CA297843.

%% file: sections/appendix.tex
%\section{Implementation details.}
%More detailed implementation settings and loss weights for the MR-to-CT task are provided in the Supplementary Material.

\section*{\centering Appendix}

\section{Additional Ablation Studies}
\label{sec:sup1}
%--------------------- maybe for supp-------
We further ablate the weights of $\mathcal{L}_{\mathrm{align}}$, $\mathcal{L}_{\mathrm{dec}}$, and $\mathcal{L}_{\mathrm{bank}}$ to study their effects on the learned shared structure space, as shown in Tabs.~\ref{tab:lambda_align}--\ref{tab:lambda_bank}. 
For the latent structure alignment loss, Tab.~\ref{tab:lambda_align} shows that $\lambda_{\mathrm{align}}=0.5$ gives the best result, achieving 19.05 dB PSNR and 64.65\% SSIM, while both smaller and larger weights degrade performance. 
This suggests that proper latent alignment is important for learning domain-shared structural representations, but excessive alignment may over-constrain the latent space. 
For the structure reconstruction loss, Tab.~\ref{tab:lambda_dec} shows that $\lambda_{\mathrm{dec}}=1$ performs best, indicating that decoding the shared latent representation into structure masks provides useful anatomical supervision. 
However, increasing the weight to 2 reduces performance, suggesting that overly strong reconstruction supervision may limit the flexibility of the structure latent. 
For the structure-bank consistency loss, Tab.~\ref{tab:lambda_bank} shows that $\lambda_{\mathrm{bank}}=0.2$ achieves the best performance. 
A smaller weight provides insufficient target-domain structural guidance, whereas a larger weight may introduce less relevant prototype constraints. 
Overall, these results demonstrate that the three Stage-I losses are complementary and require balanced weighting to learn an effective shared structure representation for downstream OCT-to-OCTA translation.

Table~\ref{tab:ablation_mask_cross} studies the effects of the Stage I coarse source mask $m_A$ and the Stage II cross-domain branch. Removing both components leads to the lowest performance, with 16.70 dB PSNR and 61.29\% SSIM on the whole image. Adding only the cross-domain branch substantially improves the whole-image performance to 18.23 dB PSNR and 63.24\% SSIM, suggesting that Stage II cross-domain guidance helps the model better match target-domain structures. Using only the coarse mask $m_A$ also improves the result to 17.82 dB PSNR and 62.96\% SSIM, showing that the Stage I source structural prior contributes to anatomical preservation. Combining both components achieves the best performance across all evaluated regions, including the whole image, vessel region, and FAZ region. In particular, the full model reaches 19.05 dB PSNR and 64.65\% SSIM on the whole image, 18.92 dB PSNR and 75.72\% SSIM on vessels, and 20.48 dB PSNR and 35.05\% SSIM on FAZ. These results indicate that the coarse source mask and cross-domain branch provide complementary structural guidance for OCT-to-OCTA translation.

We further analyze the design choice of performing structure conditioning with the memory bank in the shared latent space. As shown in Fig.~\ref{fig:prototype_structure_condition}, the decoded masks of the top-$K$ retrieved prototypes provide relevant structural candidates, but they are not perfectly aligned in the image domain. Therefore, directly aggregating these decoded masks with similarity weights leads to an unreasonable structure, as shown in the last column: the direct combination produces noisy and overly dense vessel responses that do not preserve a plausible anatomical layout. In contrast, our method first performs similarity-weighted aggregation over the retrieved prototypes in the shared structure latent space and then decodes the aggregated representation into the estimated structure mask. This latent-space aggregation yields a more reasonable and coherent structure that is closer to the target mask. The comparison indicates that the shared structure latent space provides a better manifold for combining anatomical priors, while direct image-domain aggregation fails to respect spatial and topological consistency. This supports our design of using a memory bank together with latent-space structure conditioning for faithful cross-domain translation.
%--------------------- maybe for supp-------

\section{Implementation details}
\label{sec:sup2}
\textbf{OCT-to-OCTA experiments:} In Stage I, the shared structure encoder-decoder is trained progressively: we first warm up the model with target masks $m_B$ for $10$ epochs, then introduce annotated target images $x_B$ for another $10$ epochs, and finally add source images $x_A$ together with the SMB. The Stage I loss weights are set to 
$\lambda_{\mathrm{dec}}=1.0$, 
$\lambda_{\mathrm{align}}=0.5$, 
$\lambda_{\mathrm{bank}}=0.2$, and 
$\lambda_{\mathrm{cycle}}=0.5$. 
In Stage II, after freezing the structure encoder-decoder, the conditional flow generator is trained with 
$\lambda_{\mathrm{self}}=1.0$, 
$\lambda_{\mathrm{cross}}=0.3$, and 
$\lambda_{\mathrm{struct}}=0.1$. 
All weights are kept fixed throughout training. 
For consistency across the two stages, we retrieve the top-$K$ prototypes from the memory bank with $K=4$. 

%%%%%%%%%%%%%%%Table 8%%%%%%%%%%%%%%%%%%%%
\input{tables/loss_align}
%%%%%%%%%%%%%%%Table 8%%%%%%%%%%%%%%%%%%%

%%%%%%%%%%%%%%%Table 9%%%%%%%%%%%%%%%%%%%%
\input{tables/loss_dec}
%%%%%%%%%%%%%%%Table 9%%%%%%%%%%%%%%%%%%%

%%%%%%%%%%%%%%%Table 11%%%%%%%%%%%%%%%%%%%%
\input{tables/loss_bank}
%%%%%%%%%%%%%%%Table 11%%%%%%%%%%%%%%%%%%%

%%%%%%%%%%%%%%%Table 3%%%%%%%%%%%%%%%%%%%%
\input{tables/ablation}
%%%%%%%%%%%%%%%Table 3%%%%%%%%%%%%%%%%%%%%

%%%%%%%%%%%%%%%Figure 5 9%%%%%%%%%%%%%%%%%%%%
\input{figures/inter_res}
%%%%%%%%%%%%%%%Figure 5  9%%%%%%%%%%%%%%%%%%%

\textbf{Brain MR-to-CT experiments:} We follow the same two-stage training strategy as described in the main paper, while using task-specific loss weights. 
In Stage I, the shared structure encoder-decoder is trained with 
$\lambda_{\mathrm{dec}}=1.0$, 
$\lambda_{\mathrm{align}}=0.5$, 
$\lambda_{\mathrm{bank}}=0.2$, and 
$\lambda_{\mathrm{cycle}}=0.5$.
In Stage II, the structure encoder-decoder is frozen, and the conditional flow generator is optimized with 
$\lambda_{\mathrm{self}}=1.0$, 
$\lambda_{\mathrm{cross}}=1.0$, and 
$\lambda_{\mathrm{struct}}=0.5$. 
For consistency across the two stages, we retrieve the top-$K$ prototypes from the memory bank with $K=4$. 
For the similarity weighting in the memory bank, we use the same parameters for both OCT-to-OCTA and brain MR-to-CT experiments, setting $\gamma=2$ and $\tau=0.05$. 
%%%%%%%%%%%%%%%%%%%%%% Details %%%%%%%%%%%%%%%

\myparagraph{Training procedure.} During training, we first warm up the structure encoder-decoder using only target masks $m_B$ to construct a mask-defined structural latent space. We then introduce target images $x_B$ with annotations $m_B$ to align image-derived latents with this space. Finally, we add source images $x_A$ and construct the SMB from target images and masks to weakly align source structures with the target structural distribution. After Stage I, the structure encoder-decoder is frozen to provide structural conditions for the target-domain flow generator. Before training the structure-conditioned generator in Stage II, we pretrain an unconditional pixel-space flow model using only target-domain images. This model learns the marginal appearance distribution of the target modality and serves as the initialization for the conditional flow generator. With this target-domain appearance prior, the conditional generator does not need to learn target appearance from scratch, and can focus on rendering target-domain images under the structural conditions provided by the frozen structure encoder-decoder. During Stage II, the self-branch and cross-branch are trained jointly, where the self-branch learns target-domain reconstruction under target structural conditions, and the cross-branch encourages source-to-target generation using source-derived structural conditions within the same conditional flow model.

\myparagraph{Construction of coarse source masks.}
In the source domain, the coarse mask $m_A$ is not assumed to be
a manual annotation, and its construction depends on the translation
task.
The purpose of $m_A$ is to provide a weak structural cue for source
images, rather than an accurate pixel-level segmentation target.
For OCT-to-OCTA translation, we obtain coarse OCT structural masks
by applying intensity-based thresholding to highlight major
vessel-like responses and the FAZ region, followed by basic
morphological operations to remove small noisy components and
improve spatial continuity.
These masks are therefore expected to be incomplete and noisy
compared with the target-domain OCTA structural masks.
For MR-to-CT translation, source-domain structural masks are
constructed from anatomical segmentation labels, with the label
sources and class definitions detailed in
Appendix~\ref{sec:sup_data}.
For MR subjects with paired CT, the bone label is derived from
the corresponding CT skull segmentation.
These coarse masks may contain missing regions or inaccurate
boundaries.
Our method treats them only as optional weak supervision;
the core training objective relies on the target-domain structural
masks $m_B$ to define the shared structure space.
%%%%%%%%%%%%%%%%%%%%%% Details %%%%%%%%%%%%%%%

\section{Dataset details}
\label{sec:sup_data}
%%%%%%%%%%%%%%%%%%%%%% Details %%%%%%%%%%%%%%%
For OCTA-500~\cite{li2024octa}, the original 2D OCTA projection images contain two in-plane resolutions, $304 \times 304$ and $400 \times 400$ pixels. We used the official layer-wise projections defined by the internal limiting membrane (ILM), outer plexiform layer (OPL), and Bruch's membrane (BM). Specifically, the full, ILM-OPL, and OPL-BM projections were used in Stage 1 as natural augmentation, while only the full projection was used in Stage 2. We randomly selected 50 samples as the test set and used the remaining 450 samples for training. Therefore, Stage 1 contains 1,350 training images and 150 test images, and Stage 2 contains 450 training images and 50 test images. The official OCTA-500 segmentation annotations were used, and we focused on the large-vessel and foveal avascular zone (FAZ) regions. All projection images and their corresponding masks were resized with aspect-ratio preservation and padded to $256 \times 256$.

For SynthRAD2023~\cite{thummerer2023synthrad2023}, we used the Task 1 brain subset for MRI-to-CT synthesis. The released data contain 180 paired MR/CT training subjects and 30 MR-only validation subjects. Since the CT images are unavailable for the official validation subjects, we randomly selected 20 paired training subjects as an internal test set and used the remaining paired subjects for training. The official preprocessing resampled the brain CT volumes to isotropic $1 \times 1 \times 1~\mathrm{mm}^3$ spacing and rigidly registered the MR volumes to the resampled CT space. The released volumes have case-dependent matrix sizes due to subject-specific cropping, with an average training-set size of approximately $213 \times 243 \times 194$ voxels. From these 3D volumes, we extracted 2D axial slices from the supratentorial brain and skull regions, resulting in 9,646 slices in total, including 8,685 training slices and 961 test slices. The original axial slices therefore have variable in-plane sizes, with an average size of approximately $213 \times 243$ pixels. All slices and their corresponding masks were resized with aspect-ratio preservation and padded to $256 \times 256$. CT-side anatomical labels were generated using
TotalSegmentator~\cite{wasserthal2023totalsegmentator}
and merged into six foreground classes plus background. No additional registration was applied beyond the official SynthRAD2023 preprocessing.

For Stage I training on the OCTA-500 dataset under the unpaired setting, we split the data into 225 OCT cases and 225 non-overlapping OCTA cases, ensuring no sample overlap between the two domains. For Stage II, since the cross-domain branch serves as an auxiliary augmentation to the main target-domain training, we used 100 OCT cases and 350 non-overlapping OCTA cases. For the SynthRAD2023 dataset, we used 90 CT cases and 100 MR cases in Stage I, and 140 CT cases and 50 MR cases in Stage II. MR cases from the official SynthRAD2023 validation subjects were included in both Stage I and Stage II MR sets because they do not have paired CT scans.

During the conversion from 3D volumes to 2D axial slices, we restricted the field of view to the supratentorial brain and skull region. Slices containing inferior or non-target structures, including the cerebellum, brainstem, orbits, nasal cavity, and neck, were excluded. Specifically, the lower boundary was determined by the highest axial slice containing the cerebellum plus 21 slices, which removed the cerebellum and inferior anatomy with an additional 20-slice margin. The upper boundary was set to $z_{\mathrm{dim}} - 21$, removing the last 20 slices near the top of the head. As a result, the retained slices mainly covered the cerebrum, ventricles, selected deep gray matter structures, and skull.

For anatomical label generation, the original segmentation outputs
were merged into six foreground classes plus background:
cerebrum, ventricle, thalamus, caudate, lentiform nucleus, and bone,
with background treated as a separate class.
CT-side labels were obtained from TotalSegmentator, while MR-side
brain labels were generated using SynthSeg-compatible anatomical
labels.
For MR subjects with paired CT, the bone label was derived from
the corresponding CT skull segmentation and used for the MR image.
This merging strategy was designed to provide coarse anatomical
guidance rather than fine-grained neuroanatomical parcellation.

%%%%%%%%%%%%%%%%%%%%%%%%%%%%%%%%%%%%

\section{Limitations}
\label{sec:limitation}
Although we evaluate generalization to one held-out SynthRAD2023
center, this setting does not cover the full range of institutional,
scanner, protocol, and population differences.
Our experiments use 2D en-face OCT/OCTA projections and independently
translated axial MR/CT slices, without explicitly enforcing
inter-slice consistency or 3D anatomical topology.
The framework also relies on target-domain structural masks during
training, and mask errors may affect the learned representation.
Local anatomical hallucination and omission remain possible despite
the observed improvements in structural fidelity.
Broader external validation and expert assessment are needed to
establish reliability for clinical applications.

\section{Broader impacts}
\label{sec:impacts}
The proposed framework may help improve the accessibility of multi-modal medical imaging by synthesizing target-domain images from more readily available source modalities, potentially reducing acquisition burden, cost, and patient discomfort in research and clinical-support settings. 
By explicitly modeling anatomical structure, the method aims to improve structural fidelity compared with direct translation approaches, which is important for downstream analysis and visualization. However, synthesized medical images should not be used as a substitute for acquired clinical scans without rigorous validation. Potential risks include hallucinated or missing structures, dataset bias, and reduced reliability under protocol or institution shifts. Therefore, any clinical deployment should require careful multi-center evaluation, uncertainty assessment, and expert oversight.

\section{Training details}
\label{sec:computation}
Stage~I trains the structure encoder--decoder with a batch size of 32 for 10K optimization steps, taking approximately 2 hours on 4 NVIDIA RTX A6000 GPUs under mixed precision. Stage~II trains the structure-conditioned flow-matching generator for 80K steps, finishing in roughly 2 hours on the same hardware. All otherwise unspecified hyperparameters follow their default values, including the AdamW optimizer with learning rate $2\times10^{-4}$ and weight decay $10^{-4}$, gradient clipping at 1.0, and EMA for the flow model.

%%%%%%%%%%%%%%%Figure 5 9%%%%%%%%%%%%%%%%%%%%
\input{figures/supp_quality_table3}
%%%%%%%%%%%%%%%Figure 5  9%%%%%%%%%%%%%%%%%%%

\section{Additional qualitative and robustness analyses}
\label{sec:qualitative_supp}
Fig.~\ref{fig:stage1_structure_qualitative} provides qualitative analysis of the Stage-I structure encoder under different training settings. 
The full model produces more continuous and complete vessel structures than the ablated variants, especially in the local regions highlighted by red boxes. 
Compared with the setting without the coarse source mask $m_A$, our model better preserves source-side vascular details, indicating that $m_A$ provides useful auxiliary structural guidance. 
Compared with the setting without the latent cycle consistency loss $\mathcal{L}_{\mathrm{cycle}}$, our model also reduces broken or missing vessel segments, suggesting that the cycle constraint helps maintain a more consistent shared structure space.

Figs.~\ref{fig:supp_qualitative_octa} and~\ref{fig:supp_qualitative_mr_ct} show additional qualitative results for OCT-to-OCTA and MR-to-CT translation, respectively. 
For OCT-to-OCTA translation, our method preserves vessel-like anatomical structures while generating target-domain OCTA appearance, whereas baseline methods often introduce noisy responses or lose fine vascular details. 
For MR-to-CT translation, our method produces more anatomically consistent CT-like images across different brain slices, with better preservation of tissue layouts and structure boundaries. 
These results further demonstrate that the proposed structure-conditioned generation framework improves target-domain appearance synthesis while maintaining clinically meaningful anatomical structures.

%%%%%%%%%%%%%%%%%%%%%% Details %%%%%%%%%%%%%%%

%%%%%%%%%%%%%%%Figure 5 9%%%%%%%%%%%%%%%%%%%%
\input{figures/sup_res1}
%%%%%%%%%%%%%%%Figure 5  9%%%%%%%%%%%%%%%%%%%

%%%%%%%%%%%%%%%Figure 5 9%%%%%%%%%%%%%%%%%%%%
\input{figures/sup_res2}
%%%%%%%%%%%%%%%Figure 5  9%%%%%%%%%%%%%%%%%%%

\subsection{Robustness to automatically generated target masks}
\label{sec:target_mask_robustness}

We assess sensitivity to target-domain mask quality on OCT-to-OCTA
translation.
Automatic vessel and FAZ masks are extracted from target-domain
OCTA training images using fixed thresholding and morphological
post-processing, without using the official annotations during
mask generation.
We retrain the full framework from scratch using either the official
annotations or the automatic masks, keeping all other settings fixed.
Target-domain masks are used only during training.

Tab.~\ref{tab:target_mask_robustness} reports the quality of the
automatic masks and the resulting translation performance.
Despite imperfect agreement with the official annotations,
automatic masks yield similar mean PSNR and SSIM across the
whole image, vessel, and FAZ regions.
These results support robustness to the mask errors introduced
by this automatic extraction procedure on OCT-to-OCTA.

%%%%%%%%%%%%%%%Figure 5  9%%%%%%%%%%%%%%%%%%%
\input{tables/target-mask-robustness}
%%%%%%%%%%%%%%%Figure 5  9%%%%%%%%%%%%%%%%%%%
%%%%%%%%%%%%%%%Figure 5  9%%%%%%%%%%%%%%%%%%%
\input{tables/prototype-bias}
%%%%%%%%%%%%%%%Figure 5  9%%%%%%%%%%%%%%%%%%%

%%%%%%%%%%%%%%%Figure 5  9%%%%%%%%%%%%%%%%%%%
\input{tables/user-study}
%%%%%%%%%%%%%%%Figure 5  9%%%%%%%%%%%%%%%%%%%

\subsection{Memory-bank bias and subject-specific fidelity}
\label{sec:prototype_bias}

We examine whether outputs resemble retrieved training prototypes
at the expense of subject-specific anatomy.
The memory bank is used only during training; retrieval for this
analysis is performed post hoc.
For each held-out source image, we fix the retrieved training
prototypes and their weights across all compared methods.
Using the frozen vessel evaluator, we compute Own-subject Dice
against the subject-specific annotation.
Prototype Affinity is the retrieval-weighted mean Dice between
the predicted vessel mask and the individual retrieved masks.
Subject Margin is Own-subject Dice minus Prototype Affinity.
Acquired OCTA provides a reference, while MuTri provides a comparison
method that does not use the memory bank.

Tab.~\ref{tab:prototype_bias} reports results for all 50 held-out
cases and the ten cases with the lowest retrieval confidence.
Here, confidence measures source-to-bank similarity.
Our method achieves higher mean Own-subject Dice than MuTri,
with similar Prototype Affinity and a slightly larger mean
Subject Margin.
The same pattern holds in the lowest-confidence subset.
These results are consistent with preservation of subject-specific
structure at the global level, although global overlap measures
cannot exclude local retrieval-induced hallucination or omission.

\section{Blinded reader study}
\label{sec:reader_study}

We conduct a blinded reader study on 10 held-out OCT-to-OCTA cases
with five readers, including three retina specialists with
2--10 years of experience interpreting OCT/OCTA.
Each reader independently assesses anonymized outputs from SynDiff,
MuTri, and our method relative to the same source OCT and acquired OCTA.
Clinical vascular fidelity and diagnostic acceptability are scored
from 1 to 5, with higher scores indicating better quality.
Missing-vessel and unsupported-vessel severity are scored from 0 to 3,
with lower scores indicating fewer perceived structural errors.
We report mean scores with 95\% case-level bootstrap confidence
intervals, retaining all five reader ratings within each resampled case.

As shown in Tab.~\ref{tab:reader_study}, our method achieves the
highest mean vascular fidelity and diagnostic acceptability,
together with the lowest mean missing-vessel and unsupported-vessel
severity among the evaluated methods.
These results complement the independent vessel evaluation in
Tab.~\ref{tab:independent_vessel_evaluation} by providing reader-based
evidence of improved structural fidelity.
The small cohort limits the generalizability of these findings.

%% file: tables/loss_align.tex
\begin{table}
\centering
\scriptsize
\setlength{\tabcolsep}{4pt}
\renewcommand{\arraystretch}{1.1}
\caption{Ablation on the weight $\lambda_{\mathrm{align}}$ of the latent structure alignment loss $\mathcal{L}_{\mathrm{align}}$ on the OCT-to-OCTA translation task. We report PSNR$\uparrow$ and SSIM$\uparrow$ (mean $\pm$ 95\% CI). Best results are in \best{bold}.}
\label{tab:lambda_align}
\begin{tabular}{ccc}
\toprule
\textbf{$\lambda_{\mathrm{align}}$} & \textbf{PSNR}$\uparrow$ & \textbf{SSIM}$\uparrow$ \\
\midrule
0.25 & $18.35{\pm}0.64$ & $59.39{\pm}5.33$ \\
0.5  & $\textbf{19.05}{\pm}\textbf{0.72}$ & $\textbf{64.65}{\pm}\textbf{5.40}$ \\
1    & $18.27{\pm}0.56$ & $58.43{\pm}4.97$ \\
\bottomrule
\end{tabular}
\end{table}

%% file: tables/loss_dec.tex
\begin{table}
\centering
\scriptsize
\setlength{\tabcolsep}{4pt}
\renewcommand{\arraystretch}{1.1}
\caption{Ablation on the weight $\lambda_{\mathrm{dec}}$ of the structure reconstruction loss $\mathcal{L}_{\mathrm{dec}}$ on the OCT-to-OCTA translation task. We report PSNR$\uparrow$ and SSIM$\uparrow$ (mean $\pm$ 95\% CI). Best results are in \best{bold}.}
\label{tab:lambda_dec}
\begin{tabular}{ccc}
\toprule
\textbf{$\lambda_{\mathrm{dec}}$} & \textbf{PSNR}$\uparrow$ & \textbf{SSIM}$\uparrow$ \\
\midrule
0.5 & $18.40{\pm}0.57$ & $61.31{\pm}4.87$ \\
1   & $\textbf{19.05}{\pm}\textbf{0.72}$ & $\textbf{64.65}{\pm}\textbf{5.40}$ \\
2   & $18.64{\pm}0.65$ & $59.42{\pm}5.36$ \\
\bottomrule
\end{tabular}
\end{table}

%% file: tables/loss_bank.tex
\begin{table}
\centering
\scriptsize
\setlength{\tabcolsep}{4pt}
\renewcommand{\arraystretch}{1.1}
\caption{Ablation on the weight $\lambda_{\mathrm{bank}}$ of the structure-bank consistency loss $\mathcal{L}_{\mathrm{bank}}$ on the OCT-to-OCTA translation task. We report PSNR$\uparrow$ and SSIM$\uparrow$ (mean $\pm$ 95\% CI). Best results are in \best{bold}.}
\label{tab:lambda_bank}
\begin{tabular}{ccc}
\toprule
\textbf{$\lambda_{\mathrm{bank}}$} & \textbf{PSNR}$\uparrow$ & \textbf{SSIM}$\uparrow$ \\
\midrule
0.1 & $18.32{\pm}0.65$ & $58.66{\pm}5.41$ \\
0.2 & $\textbf{19.05}{\pm}\textbf{0.72}$ & $\textbf{64.65}{\pm}\textbf{5.40}$ \\
0.4 & $18.89{\pm}0.63$ & $62.36{\pm}4.95$ \\
\bottomrule
\end{tabular}
\end{table}

%% file: tables/ablation.tex
\begin{table}[t]
\centering
\scriptsize
\caption{
Ablation of coarse source mask input $m_A$ and cross-domain branch.
We report PSNR$\uparrow$ and SSIM$\uparrow$ for the whole image, vessel region, and FAZ region.
Best results are in \best{bold}.
}
\label{tab:ablation_mask_cross}

\setlength{\tabcolsep}{3pt}
\renewcommand{\arraystretch}{1.08}

\resizebox{\columnwidth}{!}{%
\begin{tabular}{@{}cc|cccccc@{}}
\toprule
\multirow{2}{*}{\textbf{$m_A$}} &
\multirow{2}{*}{\textbf{Cross branch}} &
\multicolumn{2}{c}{\textbf{Whole Image}} &
\multicolumn{2}{c}{\textbf{Vessel}} &
\multicolumn{2}{c}{\textbf{FAZ}} \\
\cmidrule(lr){3-4}
\cmidrule(lr){5-6}
\cmidrule(lr){7-8}
& &
\textbf{PSNR}$\uparrow$ & \textbf{SSIM}$\uparrow$ &
\textbf{PSNR}$\uparrow$ & \textbf{SSIM}$\uparrow$ &
\textbf{PSNR}$\uparrow$ & \textbf{SSIM}$\uparrow$ \\
\midrule
$\times$    & $\times$    & $16.70{\pm}0.74$ & $61.29{\pm}5.65$ & $16.44{\pm}0.95$ & $70.57{\pm}6.08$ & $18.19{\pm}0.99$ & $31.46{\pm}3.89$ \\
$\times$    & \checkmark  & $18.23{\pm}0.99$ & $63.24{\pm}5.59$ & $17.72{\pm}0.90$ & $74.26{\pm}5.91$ & $19.34{\pm}0.96$ & $33.36{\pm}4.00$\\
\checkmark  & $\times$    & $17.82{\pm}0.73$ & $62.96{\pm}5.26$ & $17.18{\pm}0.90$ & $72.69{\pm}5.63$ & $19.08{\pm}1.00$ & $32.10{\pm}3.88$ \\
\checkmark  & \checkmark  & $\textbf{19.05}{\pm}\textbf{0.72}$ & $\textbf{64.65}{\pm}\textbf{5.40}$ & $\textbf{18.92}{\pm}\textbf{0.88}$ & $\textbf{75.72}{\pm}\textbf{5.81}$ & $\textbf{20.48}{\pm}\textbf{1.00}$ & $\textbf{35.05}{\pm}\textbf{4.03}$ \\
\bottomrule
\end{tabular}%
}
\end{table}

%% file: figures/inter_res.tex
% \begin{wrapfigure}{r}{0.48\textwidth}
%     \vspace{-10pt}
%     \centering
%     \includegraphics[width=\linewidth]{figures/inter_res.png}
%     \caption{
%     \textbf{Latent-space structure aggregation.}
%     Direct combination obtains a structure mask by applying similarity-weighted combination to the decoded masks of the top-$K$ prototypes.
%     In contrast, our method first aggregates the top-$K$ prototypes in the shared structure latent space and then decodes the aggregated representation into $\hat{m}_A$.
%     This makes $\hat{m}_A$ closer to the true structure mask and supports our pipeline design, where the structure latent space serves as the structure condition.
%     }
%     \label{fig:prototype_structure_condition}
%     \vspace{-10pt}
% \end{wrapfigure}

\begin{figure}[t]
    \centering
    \includegraphics[width=0.75\linewidth]{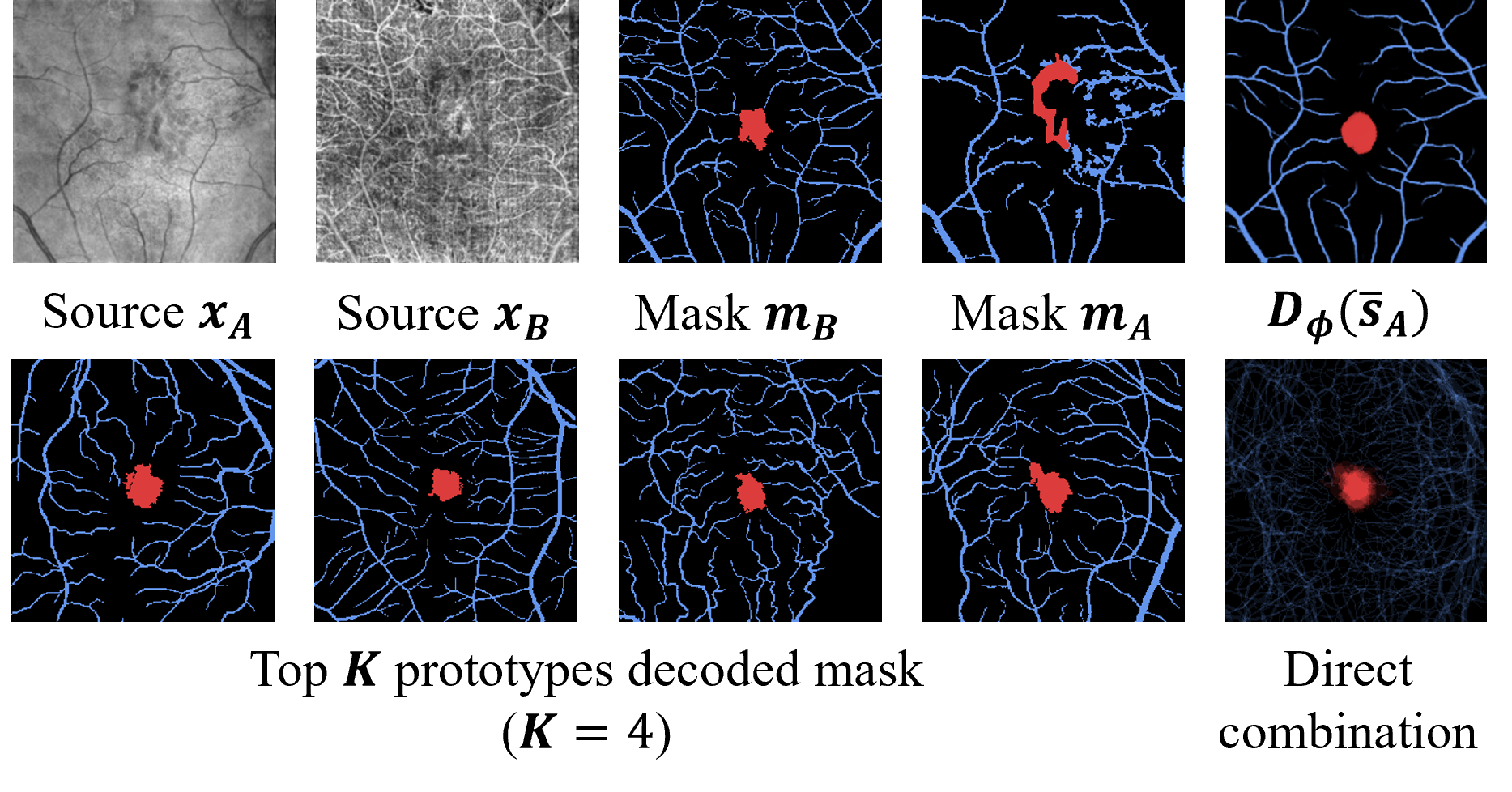}
    \caption{
\textbf{Memory bank analysis.}
Given a source image $x_A$, we retrieve the top-$K$ target-domain prototypes from the structure memory bank based on mask similarity. 
The bottom row shows the decoded masks of the retrieved prototypes, while the rightmost result shows direct similarity-weighted combination in the mask space. 
In contrast, our method first aggregates the retrieved prototypes in the shared structure latent space and then decodes the aggregated latent as $D_{\phi}(\bar{s}_A)$. 
Compared with direct mask-space combination, latent-space aggregation produces a cleaner and more coherent structural map that better matches the source-derived structure, supporting our design of using the shared structure latent space to construct structural conditions.
}
    \label{fig:prototype_structure_condition}
    \vspace{-8pt}
\end{figure}

%% file: figures/supp_quality_table3.tex
\begin{figure}
    \centering
    \includegraphics[width=\textwidth]{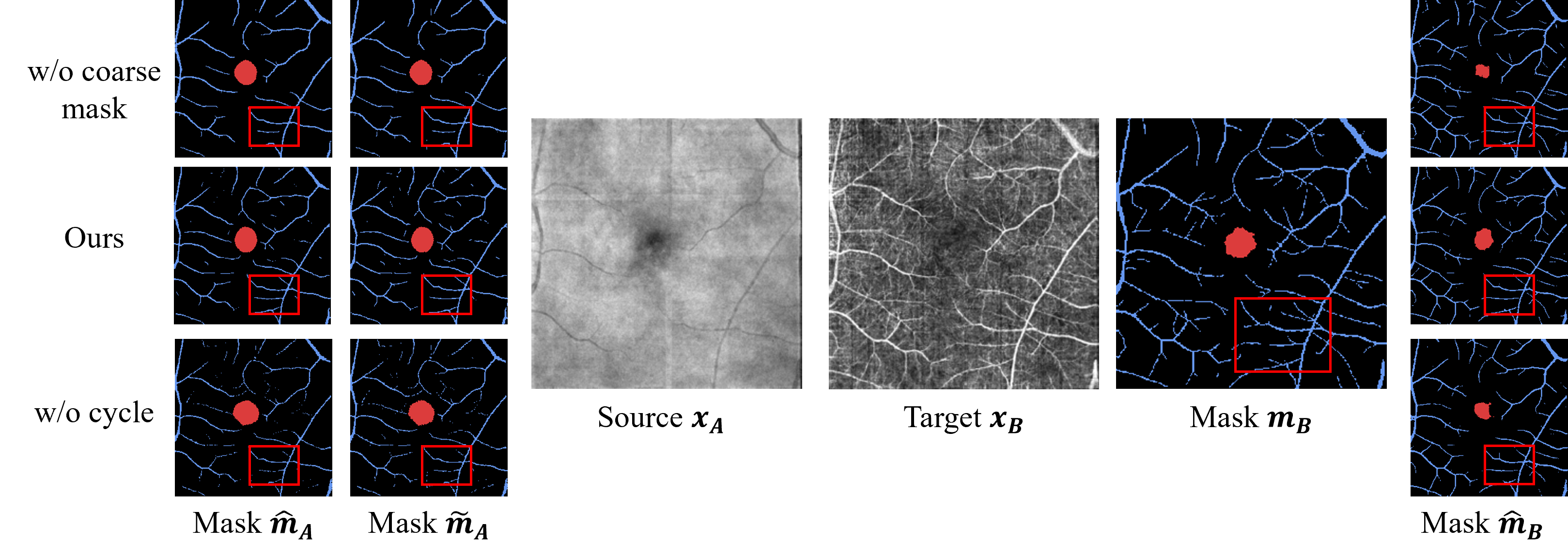}
    \caption{
    \textbf{Qualitative analysis of Stage I structure encoding.}
    We visualize the predicted structure masks under different Stage I settings, corresponding to Table~\ref{tab:structure_dice_ablation}. 
    Red boxes highlight local vascular regions. 
    Compared with the ablated variants, the full model preserves more continuous vessel structures with fewer broken or missing segments, demonstrating the benefit of using the source coarse mask $m_A$ and the latent cycle consistency loss $\mathcal{L}_{\mathrm{cycle}}$ for learning a shared structure space.
    }
    \label{fig:stage1_structure_qualitative}
\vspace{-12pt}
\end{figure}

%% file: figures/sup_res1.tex
\begin{figure}[p]
    \centering
    \includegraphics[
        width=\linewidth,
        height=0.95\textheight,
        keepaspectratio
    ]{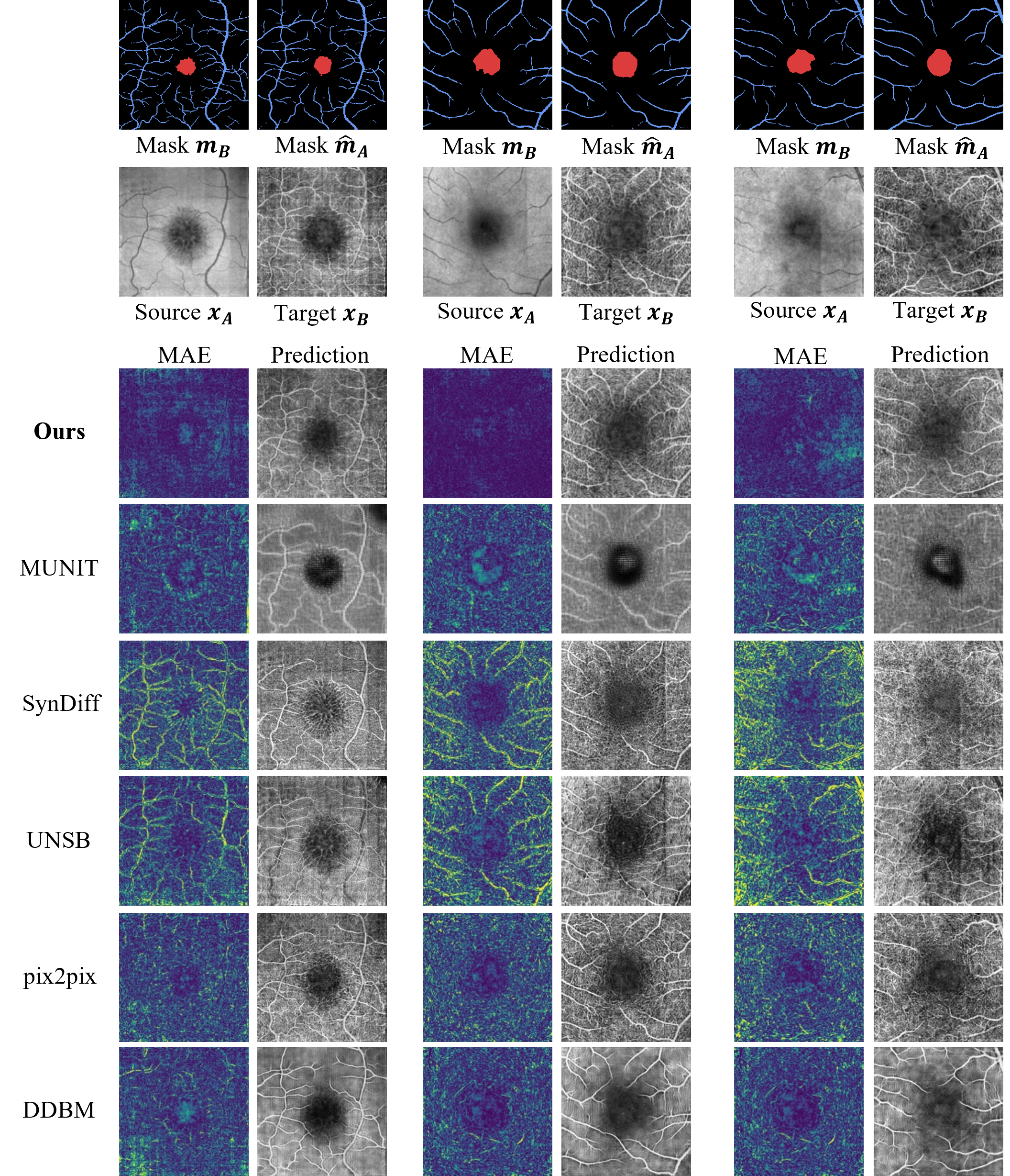}
    \caption{
    Additional qualitative results for OCT-to-OCTA translation. 
    The figure shows multiple representative examples with the corresponding structural annotations, source/target images, intermediate response maps, and generated OCTA results. 
    These examples further demonstrate that the proposed structure-conditioned generation preserves vessel-like anatomical structures while rendering target-domain OCTA appearance.
    }
    \label{fig:supp_qualitative_octa}
\end{figure}

%% file: figures/sup_res2.tex
\begin{figure}[p]
    \centering
    \includegraphics[
        width=\linewidth,
        height=0.95\textheight,
        keepaspectratio
    ]{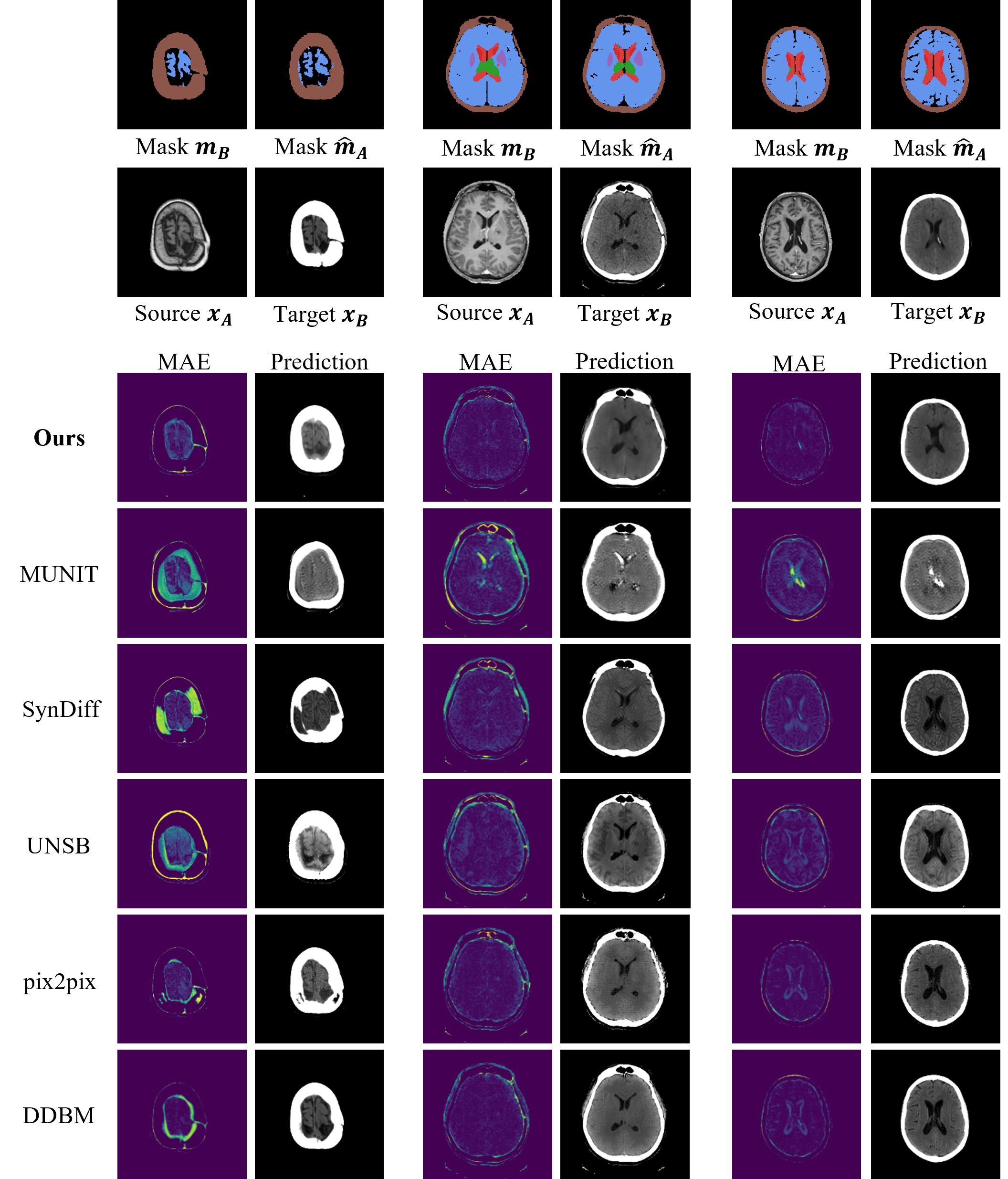}
    \caption{
    Additional qualitative results for MR-to-CT translation. 
    The figure presents representative examples with structural annotations, source MR images, target-domain CT images, intermediate response maps, and generated CT results. 
    These examples further show that the proposed structure-conditioned generation preserves anatomical structure while producing CT-like appearance across different brain slices.
    }
    \label{fig:supp_qualitative_mr_ct}
\end{figure}

%% file: tables/target-mask-robustness.tex
\begin{table}
\centering
\caption{
Robustness to automatically generated target masks on OCT-to-OCTA.
Mask Dice is measured against official annotations and reported as
mean $\pm$ standard deviation.
Translation metrics are reported as means.
PSNR is in dB; Dice and SSIM are in \%.
}
\label{tab:target_mask_robustness}
\small
\setlength{\tabcolsep}{3pt}
\renewcommand{\arraystretch}{1.08}

\resizebox{\linewidth}{!}{%
\begin{tabular}{@{}lcccccccc@{}}
\toprule
\multirow{2}{*}{\textbf{Target masks}} &
\multicolumn{2}{c}{\textbf{Mask Dice}} &
\multicolumn{2}{c}{\textbf{Whole Image}} &
\multicolumn{2}{c}{\textbf{Vessel}} &
\multicolumn{2}{c}{\textbf{FAZ}} \\
\cmidrule(lr){2-3}
\cmidrule(lr){4-5}
\cmidrule(lr){6-7}
\cmidrule(lr){8-9}
&
\textbf{Vessel} &
\textbf{FAZ} &
\textbf{PSNR} $\uparrow$ &
\textbf{SSIM} $\uparrow$ &
\textbf{PSNR} $\uparrow$ &
\textbf{SSIM} $\uparrow$ &
\textbf{PSNR} $\uparrow$ &
\textbf{SSIM} $\uparrow$ \\
\midrule

Official
& $100.00$
& $100.00$
& $19.05$
& $64.65$
& $18.92$
& $75.72$
& $20.48$
& $35.05$ \\

Automatic
& $81.65\pm3.10$
& $80.34\pm13.30$
& $19.03$
& $64.39$
& $18.95$
& $75.34$
& $20.42$
& $35.19$ \\

\bottomrule
\end{tabular}%
}
\end{table}

%% file: tables/prototype-bias.tex
\begin{table}
\centering
\caption{
Post-hoc analysis of subject-specific fidelity and prototype affinity
on OCT-to-OCTA.
All methods are evaluated against the same retrieved training
prototypes for each case.
Subject Margin is Own-subject Dice minus Prototype Affinity.
Values are mean $\pm$ 95\% case-level bootstrap CI half-width.
Bold indicates the best mean among translation methods for
Own-subject Dice and Subject Margin.
}
\label{tab:prototype_bias}
\small
\setlength{\tabcolsep}{4pt}
\renewcommand{\arraystretch}{1.08}

\resizebox{\linewidth}{!}{%
\begin{tabular}{@{}llccc@{}}
\toprule
\textbf{Subset} &
\textbf{Image / Method} &
\textbf{Own-subject Dice} $\uparrow$ &
\textbf{Prototype Affinity} &
\textbf{Subject Margin} $\uparrow$ \\
\midrule

\multirow{3}{*}{All 50}
& Acquired OCTA
& $0.859\pm0.008$
& $0.124\pm0.004$
& $0.735\pm0.009$ \\

& MuTri
& $0.787\pm0.017$
& $0.112\pm0.003$
& $0.675\pm0.017$ \\

& Ours
& $\mathbf{0.797}\pm0.026$
& $0.118\pm0.004$
& $\mathbf{0.679}\pm0.025$ \\

\midrule

\multirow{3}{*}{\shortstack[l]{Lowest retrieval-\\confidence 10}}
& Acquired OCTA
& $0.867\pm0.016$
& $0.111\pm0.006$
& $0.755\pm0.021$ \\

& MuTri
& $0.810\pm0.039$
& $0.104\pm0.005$
& $0.706\pm0.040$ \\

& Ours
& $\mathbf{0.827}\pm0.038$
& $0.107\pm0.004$
& $\mathbf{0.720}\pm0.041$ \\

\bottomrule
\end{tabular}%
}
\end{table}

%% file: tables/user-study.tex
\begin{table}
\centering
\caption{
Blinded reader study on 10 held-out OCT-to-OCTA cases with five readers.
Fidelity and Acceptability denote clinical vascular fidelity and
diagnostic acceptability; Missing and Unsupported denote missing-vessel
and unsupported-vessel severity.
Values are mean $\pm$ 95\% case-level bootstrap CI half-width.
Bold indicates the best mean.
}
\label{tab:reader_study}
\small
\setlength{\tabcolsep}{6pt}
\renewcommand{\arraystretch}{1.1}

\begin{tabular}{@{}lcccc@{}}
\toprule
\textbf{Method} &
\textbf{Fidelity} $\uparrow$ &
\textbf{Missing} $\downarrow$ &
\textbf{Unsupported} $\downarrow$ &
\textbf{Acceptability} $\uparrow$ \\
\midrule

SynDiff
& $2.74\pm0.34$
& $1.74\pm0.28$
& $1.42\pm0.24$
& $2.64\pm0.39$ \\

MuTri
& $3.00\pm0.17$
& $1.64\pm0.15$
& $1.18\pm0.16$
& $2.94\pm0.21$ \\

Ours
& $\mathbf{3.38}\pm0.60$
& $\mathbf{1.42}\pm0.49$
& $\mathbf{0.88}\pm0.29$
& $\mathbf{3.38}\pm0.75$ \\

\bottomrule
\end{tabular}
\end{table}

%% file: reference.bib
@String(CVPR= {IEEE Conf. Comput. Vis. Pattern Recog.})

@String(ECCV= {Eur. Conf. Comput. Vis.})

@String(CVPR  = {CVPR})

@String(ECCV  = {ECCV})

@inproceedings{zhu2017unpaired,
  title={Unpaired image-to-image translation using cycle-consistent adversarial networks},
  author={Zhu, Jun-Yan and Park, Taesung and Isola, Phillip and Efros, Alexei A},
  booktitle={Proceedings of the IEEE international conference on computer vision},
  pages={2223--2232},
  year={2017}
}

@inproceedings{park2020contrastive,
  title={Contrastive learning for unpaired image-to-image translation},
  author={Park, Taesung and Efros, Alexei A and Zhang, Richard and Zhu, Jun-Yan},
  booktitle={European conference on computer vision},
  pages={319--345},
  year={2020},
  organization={Springer}
}

@article{syndiff,
  title={Unsupervised medical image translation with adversarial diffusion models},
  author={{\"O}zbey, Muzaffer and Dalmaz, Onat and Dar, Salman UH and Bedel, Hasan A and {\"O}zturk, {\c{S}}aban and G{\"u}ng{\"o}r, Alper and Cukur, Tolga},
  journal={IEEE Transactions on Medical Imaging},
  volume={42},
  number={12},
  pages={3524--3539},
  year={2023},
  publisher={IEEE}
}

@article{lipman2022flow,
  title={Flow matching for generative modeling},
  author={Lipman, Yaron and Chen, Ricky TQ and Ben-Hamu, Heli and Nickel, Maximilian and Le, Matt},
  journal={arXiv preprint arXiv:2210.02747},
  year={2022}
}

@inproceedings{isola2017image,
  title={Image-to-image translation with conditional adversarial networks},
  author={Isola, Phillip and Zhu, Jun-Yan and Zhou, Tinghui and Efros, Alexei A},
  booktitle={Proceedings of the IEEE conference on computer vision and pattern recognition},
  pages={1125--1134},
  year={2017}
}

@article{zhou2023denoising,
  title={Denoising diffusion bridge models},
  author={Zhou, Linqi and Lou, Aaron and Khanna, Samar and Ermon, Stefano},
  journal={arXiv preprint arXiv:2309.16948},
  year={2023}
}

@inproceedings{chen2025mutri,
  title={Mutri: Multi-view tri-alignment for oct to octa 3d image translation},
  author={Chen, Zhuangzhuang and Wang, Hualiang and Ou, Chubin and Li, Xiaomeng},
  booktitle={Proceedings of the Computer Vision and Pattern Recognition Conference},
  pages={20885--20894},
  year={2025}
}

@inproceedings{zhang2025structure,
  title={Structure-Aware MRI Translation: Multi-modal Latent Diffusion Model with Arbitrary Missing Modalities},
  author={Zhang, Xinzhe and Liang, Junjie and Cao, Peng and Yang, Jinzhu and Zaiane, Osmar R},
  booktitle={International Conference on Medical Image Computing and Computer-Assisted Intervention},
  pages={508--518},
  year={2025},
  organization={Springer}
}

@inproceedings{huang2018multimodal,
  title={Multimodal unsupervised image-to-image translation},
  author={Huang, Xun and Liu, Ming-Yu and Belongie, Serge and Kautz, Jan},
  booktitle={Proceedings of the European conference on computer vision (ECCV)},
  pages={172--189},
  year={2018}
}

@inproceedings{kim2024unpaired,
  title={Unpaired image-to-image translation via neural schr{\"o}dinger bridge},
  author={Kim, Beomsu and Kwon, Gihyun and Kim, Kwanyoung and Ye, Jong Chul},
  booktitle={International Conference on Learning Representations},
  volume={2024},
  pages={19312--19331},
  year={2024}
}

@article{ho2020denoising,
  title={Denoising diffusion probabilistic models},
  author={Ho, Jonathan and Jain, Ajay and Abbeel, Pieter},
  journal={Advances in neural information processing systems},
  volume={33},
  pages={6840--6851},
  year={2020}
}

@article{zhang2024lefusion,
  title={Lefusion: Controllable pathology synthesis via lesion-focused diffusion models},
  author={Zhang, Hantao and Liu, Yuhe and Yang, Jiancheng and Wan, Shouhong and Wang, Xinyuan and Peng, Wei and Fua, Pascal},
  journal={arXiv preprint arXiv:2403.14066},
  year={2024}
}

@article{hansen2024inpainting,
  title={Inpainting pathology in lumbar spine MRI with latent diffusion},
  author={Hansen, Colin and Glinskis, Simas and Raju, Ashwin and Kornreich, Micha and Park, JinHyeong and Pawar, Jayashri and Herzog, Richard and Zhang, Li and Odry, Benjamin},
  journal={arXiv preprint arXiv:2406.02477},
  year={2024}
}

@article{ahmed2025pinn,
  title={PINN-DADif: Physics-informed deep adaptive diffusion network for robust and efficient MRI reconstruction},
  author={Ahmed, Shahzad and Feng, Jinchao and Mehmood, Atif and Ali, Muhammad Usman and Yaqub, Muhammad and Manan, Malik Abdul and Raheem, Abdul},
  journal={Digital Signal Processing},
  volume={160},
  pages={105085},
  year={2025},
  publisher={Elsevier}
}

@article{xie2024dose,
  title={Dose-aware diffusion model for 3D low-dose PET: multi-institutional validation with reader study and real low-dose data},
  author={Xie, Huidong and Gan, Weijie and Zhou, Bo and Chen, Ming-Kai and Kulon, Michal and Boustani, Annemarie and Spencer, Benjamin A and Bayerlein, Reimund and Chen, Xiongchao and Liu, Qiong and others},
  journal={arXiv preprint arXiv:2405.12996},
  year={2024},
  publisher={Sept}
}

@article{hein2024ppfm,
  title={PPFM: Image denoising in photon-counting CT using single-step posterior sampling Poisson flow generative models},
  author={Hein, Dennis and Holmin, Staffan and Szczykutowicz, Timothy and Maltz, Jonathan S and Danielsson, Mats and Wang, Ge and Persson, Mats},
  journal={IEEE Transactions on Radiation and Plasma Medical Sciences},
  volume={8},
  number={7},
  pages={788--799},
  year={2024},
  publisher={IEEE}
}

@article{gupta2024topodiffusionnet,
  title={Topodiffusionnet: A topology-aware diffusion model},
  author={Gupta, Saumya and Samaras, Dimitris and Chen, Chao},
  journal={arXiv preprint arXiv:2410.16646},
  year={2024}
}

@inproceedings{xu2025topocellgen,
  title={Topocellgen: Generating histopathology cell topology with a diffusion model},
  author={Xu, Meilong and Gupta, Saumya and Hu, Xiaoling and Li, Chen and Abousamra, Shahira and Samaras, Dimitris and Prasanna, Prateek and Chen, Chao},
  booktitle={Proceedings of the Computer Vision and Pattern Recognition Conference},
  pages={20979--20989},
  year={2025}
}

@article{asadi2026fms,
  title={{FMS}$^2$: Unified Flow Matching for Segmentation and Synthesis of Thin Structures},
  author={Asadi, Babak and Wu, Peiyang and Golparvar-Fard, Mani and Shah, Viraj and Hajj, Ramez},
  journal={arXiv preprint arXiv:2603.13659},
  year={2026}
}

@article{li2024octa,
  title={OCTA-500: a retinal dataset for optical coherence tomography angiography study},
  author={Li, Mingchao and Huang, Kun and Xu, Qiuzhuo and Yang, Jiadong and Zhang, Yuhan and Ji, Zexuan and Xie, Keren and Yuan, Songtao and Liu, Qinghuai and Chen, Qiang},
  journal={Medical image analysis},
  volume={93},
  pages={103092},
  year={2024},
  publisher={Elsevier}
}

@article{thummerer2023synthrad2023,
  title={SynthRAD2023 Grand Challenge dataset: Generating synthetic CT for radiotherapy},
  author={Thummerer, Adrian and Van der Bijl, Erik and Galapon Jr, Arthur and Verhoeff, Joost JC and Langendijk, Johannes A and Both, Stefan and van den Berg, Cornelis (Nico) AT and Maspero, Matteo},
  journal={Medical physics},
  volume={50},
  number={7},
  pages={4664--4674},
  year={2023},
  publisher={Wiley Online Library}
}

@article{wasserthal2023totalsegmentator,
  title={TotalSegmentator: robust segmentation of 104 anatomic structures in CT images},
  author={Wasserthal, Jakob and Breit, Hanns-Christian and Meyer, Manfred T and Pradella, Maurice and Hinck, Daniel and Sauter, Alexander W and Heye, Tobias and Boll, Daniel T and Cyriac, Joshy and Yang, Shan and others},
  journal={Radiology: Artificial Intelligence},
  volume={5},
  number={5},
  pages={e230024},
  year={2023},
  publisher={Radiological Society of North America}
}

@article{van2017neural,
  title={Neural discrete representation learning},
  author={Van Den Oord, Aaron and Vinyals, Oriol and others},
  journal={Advances in neural information processing systems},
  volume={30},
  year={2017}
}

@article{wang2004image,
  title={Image quality assessment: from error visibility to structural similarity},
  author={Wang, Zhou and Bovik, Alan C and Sheikh, Hamid R and Simoncelli, Eero P},
  journal={IEEE transactions on image processing},
  volume={13},
  number={4},
  pages={600--612},
  year={2004},
  publisher={IEEE}
}

@inproceedings{hore2010image,
  title={Image quality metrics: PSNR vs. SSIM},
  author={Hore, Alain and Ziou, Djemel},
  booktitle={2010 20th international conference on pattern recognition},
  pages={2366--2369},
  year={2010},
  organization={IEEE}
}

@article{dice1945measures,
  title={Measures of the amount of ecologic association between species},
  author={Dice, Lee R},
  journal={Ecology},
  volume={26},
  number={3},
  pages={297--302},
  year={1945},
  publisher={JSTOR}
}

@inproceedings{fu2019geometry,
  title={Geometry-consistent generative adversarial networks for one-sided unsupervised domain mapping},
  author={Fu, Huan and Gong, Mingming and Wang, Chaohui and Batmanghelich, Kayhan and Zhang, Kun and Tao, Dacheng},
  booktitle={Proceedings of the IEEE/CVF conference on computer vision and pattern recognition},
  pages={2427--2436},
  year={2019}
}

@article{benaim2017one,
  title={One-sided unsupervised domain mapping},
  author={Benaim, Sagie and Wolf, Lior},
  journal={Advances in neural information processing systems},
  volume={30},
  year={2017}
}

@inproceedings{jung2022exploring,
  title={Exploring patch-wise semantic relation for contrastive learning in image-to-image translation tasks},
  author={Jung, Chanyong and Kwon, Gihyun and Ye, Jong Chul},
  booktitle={Proceedings of the IEEE/CVF conference on computer vision and pattern recognition},
  pages={18260--18269},
  year={2022}
}

@inproceedings{wang2021instance,
  title={Instance-wise hard negative example generation for contrastive learning in unpaired image-to-image translation},
  author={Wang, Weilun and Zhou, Wengang and Bao, Jianmin and Chen, Dong and Li, Houqiang},
  booktitle={Proceedings of the IEEE/CVF international conference on computer vision},
  pages={14020--14029},
  year={2021}
}

@inproceedings{zheng2021spatially,
  title={The spatially-correlative loss for various image translation tasks},
  author={Zheng, Chuanxia and Cham, Tat-Jen and Cai, Jianfei},
  booktitle={Proceedings of the IEEE/CVF conference on computer vision and pattern recognition},
  pages={16407--16417},
  year={2021}
}

@article{yi2026degradation,
  title={Degradation-Aware Dynamic Schr{\"o}dinger Bridge for Unpaired Image Restoration},
  author={Yi, Jingjun and Bi, Qi and Zheng, Hao and Huang, Huimin and Shen, Yixian and Zhan, Haolan and Ji, Wei and Huang, Yawen and Li, Yuexiang and Wu, Xian and others},
  journal={Advances in Neural Information Processing Systems},
  volume={38},
  pages={2415--2439},
  year={2026}
}

@inproceedings{shit2021cldice,
  title={clDice-a novel topology-preserving loss function for tubular structure segmentation},
  author={Shit, Suprosanna and Paetzold, Johannes C and Sekuboyina, Anjany and Ezhov, Ivan and Unger, Alexander and Zhylka, Andrey and Pluim, Josien PW and Bauer, Ulrich and Menze, Bjoern H},
  booktitle={2021 IEEE/CVF Conference on Computer Vision and Pattern Recognition (CVPR)},
  pages={16555--16564},
  year={2021},
  organization={IEEE}
}
